\documentclass[journal]{template/IEEEtran}

\usepackage{graphicx}
\usepackage{amsmath,amssymb,enumerate}

\usepackage{amsthm}
\usepackage{setspace}
\usepackage{booktabs}
\usepackage[usenames,dvipsnames,svgnames,table]{xcolor}
\usepackage{mathtools}
\usepackage{algorithm, algorithmicx, algpseudocode}

\usepackage{blindtext}
\usepackage{gensymb}
\usepackage{xparse}
\usepackage{lipsum}
\usepackage{mathrsfs}
\usepackage[mathscr]{euscript}
\usepackage{times}
\usepackage[noadjust]{cite} %
\usepackage[numbers]{natbib}
\usepackage{multicol}
\usepackage{multirow}
\definecolor{darkgreen}{rgb}{0,0.6,0}
\usepackage[bookmarks=true,colorlinks=true,pdfpagemode=UseNone,citecolor=black,linkcolor=black,urlcolor=BrickRed,pagebackref]{hyperref}
\usepackage[caption=false,font=footnotesize]{subfig}
\usepackage{amsfonts}
\usepackage[utf8]{inputenc}
\usepackage[T1]{fontenc}
\usepackage{textcomp}
\usepackage{arydshln}
\usepackage{tabu}
\usepackage{cuted}

\newtheorem{remark}{Remark}

\definecolor{note}{rgb}{0.1,0.1,1}
\definecolor{rephase}{rgb}{0.15,0.7,0.15}
\definecolor{bag}{rgb}{0.6,0.6,0.2}

\makeatletter
\renewcommand*\env@matrix[1][c]{\hskip -\arraycolsep
  \let\@ifnextchar\new@ifnextchar
  \array{*\c@MaxMatrixCols #1}}
\makeatother

\newcommand{\m}{\mathop{\mathrm{m}}}

\DeclareDocumentCommand{\vector}{ O{} }{\mathrm{vec}(#1)}

\makeatletter
\newcommand{\mathleft}{\@fleqntrue\@mathmargin0pt}
\newcommand{\mathcenter}{\@fleqnfalse}
\makeatother

\usepackage{booktabs}
\usepackage[normalem]{ulem}
\newcommand\redout{\bgroup\markoverwith
{\textcolor{orange}{\rule[.5ex]{2pt}{1.5pt}}}\ULon}

\newcommand{\bl}[1]{{\textcolor{black}{#1}}}

\definecolor{jgreen}{RGB}{0,155,0}

\providecommand{\secref}[1]{\mbox{Section~\ref{#1}}}

\begin{document}
\title{Multitask Learning for Scalable and Dense Multilayer Bayesian Map Inference}

\author{Lu Gan, Youngji Kim, Jessy W. Grizzle, Jeffrey M. Walls, Ayoung Kim, Ryan M. Eustice, and Maani Ghaffari%
\thanks{Toyota Research Institute provided funds to support this work. Funding for M. Ghaffari was in part provided by NSF Award No. 2118818. Funding for J. Grizzle was in part provided by NSF Award No. 2118818. Y.~Kim was supported by the International Research \& Development Program of the National Research Foundation of Korea (NRF) funded by the Ministry of Science and ICT (Grant number: 2019K1A3A1A12069741).~\textit{(Lu Gan and Youngji Kim contributed equally to this work.)}~\textit{(Corresponding author: Lu Gan.)}}
\thanks{L. Gan, J. Grizzle, R. Eustice, and M. Ghaffari are with the Robotics Institute, University of Michigan, Ann Arbor, MI 48109, USA.~\texttt{\{ganlu, grizzle, eustice, maanigj\}@umich.edu}.}%
\thanks{Y. Kim is with the Department of Civil and Environmental Engineering, KAIST, Daejeon, S. Korea.~\texttt{youngjikim@kaist.ac.kr}.}%
\thanks{A. Kim is with the Department of Mechanical Engineering, SNU, Seoul, S. Korea.~\texttt{ayoungk@snu.ac.kr}.}%
\thanks{J. Walls is with Woven Planet Holdings, Inc.~\texttt{jeff.walls@woven-planet.global}.}
}

\maketitle

\begin{abstract}
This article presents a novel and flexible multitask multilayer Bayesian mapping framework with readily extendable attribute layers. The proposed framework goes beyond modern metric-semantic maps to provide even richer environmental information for robots in a single mapping formalism while exploiting intralayer and interlayer correlations. It removes the need for a robot to access and process information from many separate maps when performing a complex task, advancing the way robots interact with their environments. To this end, we design a multitask deep neural network with attention mechanisms as our front-end to provide heterogeneous observations for multiple map layers simultaneously. Our back-end runs a scalable closed-form Bayesian inference with only logarithmic time complexity. 
We apply the framework to build a dense robotic map including metric-semantic occupancy and traversability layers. Traversability ground truth labels are automatically generated from exteroceptive sensory data in a self-supervised manner. We present extensive experimental results on publicly available datasets and data collected by a 3D bipedal robot platform and show reliable mapping performance in different environments. Finally, we also discuss how the current framework can be extended to incorporate more information such as friction, signal strength, temperature, and physical quantity concentration using Gaussian map layers. The software for reproducing the presented results or running on customized data is made publicly available. 
\end{abstract}


\IEEEpeerreviewmaketitle

\section{Introduction}

\IEEEPARstart{R}{obotic} mapping is the process of inferring a model of the environment from noisy measurements and is an essential task in the pursuit of robotic autonomy~\cite{thrun2002robotic}. Over the recent decades of highly active research on robotic mapping, maps have used different representations, included different sensing modalities, and been applied to a variety of tasks such as localization~\cite{dube2020segmap}, as a reference for navigation~\cite{kim2013perception}, and autonomous exploration~\cite{jadidi2018gaussian}. Traditionally, maps built for robot localization contain geometric information of the environment. For example, occupancy grid maps employ voxel grids as the map representation and probabilistically assign a binary label to each grid cell to denote whether it is occupied~\cite{elfes1989using, hornung2013octomap, doherty2019learning}. The geometric structure of the scene is also encoded in dense maps where Truncated Signed Distance Fields (TSDFs) are used to represent a surface implicitly~\cite{newcombe2011kinectfusion, oleynikova2017voxblox}. For better visualization~\cite{bylow2013real} and camera tracking~\cite{henry2014rgb}, 3D maps are also textured with appearance (e.g., color) information.

Although classical robotic maps have reached a level of maturity for localization purposes, most practical robotic applications require more than just geometric information, e.g., high-level path planning and task planning. Semantic mapping timely extends the map representation to include semantic knowledge in addition to geometry and appearance, and becomes a highly active research area~\cite{wolf2008semantic, ghaffari2017gaussian, gan2017sparse, mccormac2017semanticfusion, rosinol2020kimera, hiller2019learning, zobeidi2020dense, gan2020bayesian}. In recent years, object instances, as another form of semantic knowledge, are also incorporated into robotic maps~\cite{salas2013slam++, sunderhauf2017meaningful, grinvald2019volumetric}. 

\begin{figure}[t]
    \centering
    \includegraphics[width=0.98\columnwidth,  trim={0cm 17cm 22cm 0cm}, clip]{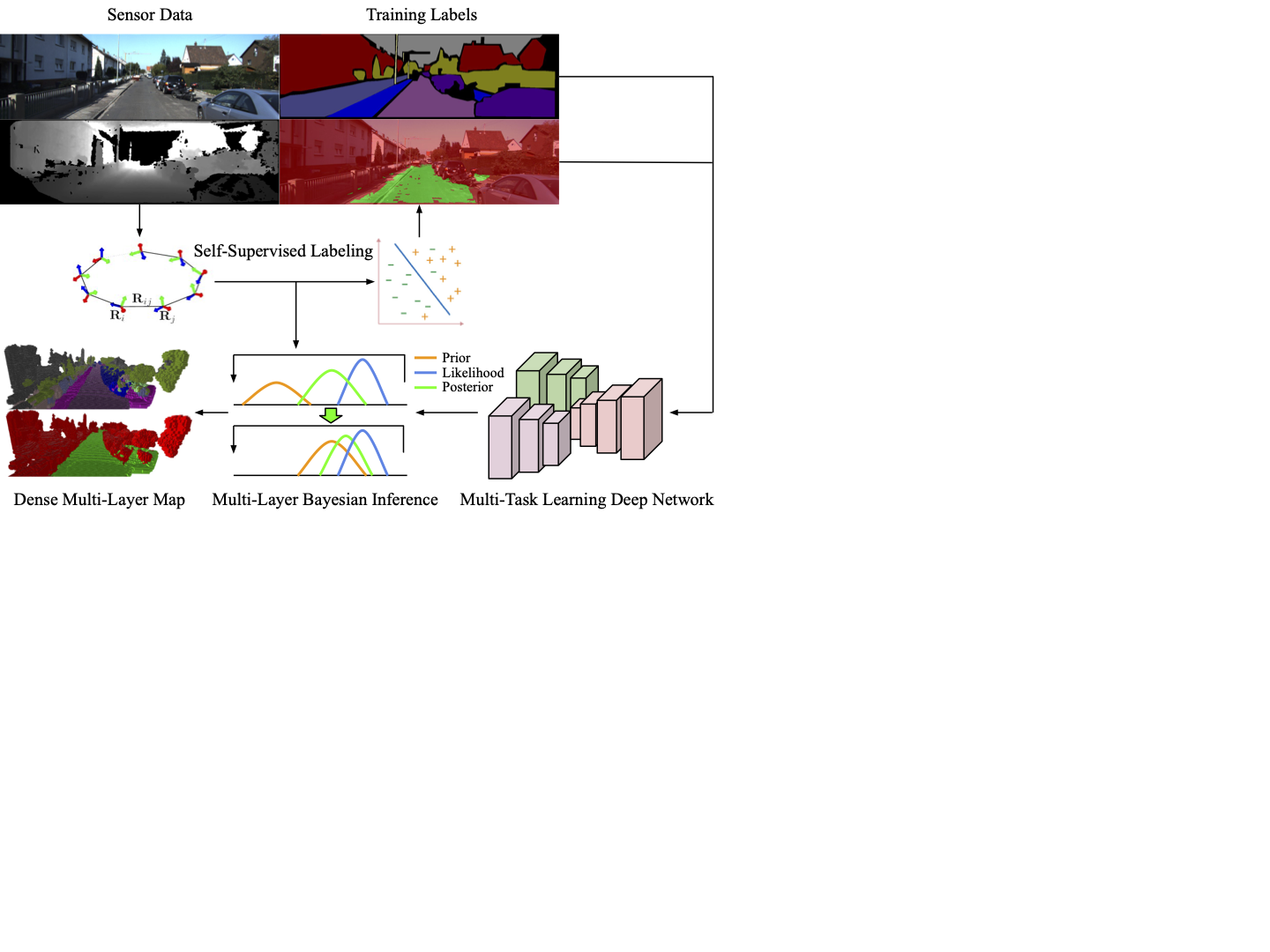}
    \caption{Overview of our proposed multitask multilayer Bayesian mapping framework. We first employ exteroceptive sensory data and sensor trajectory to generate training data for map attributes other than semantics (traversability specifically) in a self-supervised manner. We then train a unified segmentation deep neural network for monocular image via MTL using available data. Predictions of the MTL network are processed through our multilayer Bayesian inference sequentially and efficiently to build dense maps. Inter-layer correlations are also leveraged to improve the map posterior. Our framework is designed to be easily extended to include additional layers.}
    \label{fig:overview}
\end{figure}

With the inclusion of scene understanding, semantic maps enable a much greater range of tasks for robots. For instance, domestic robots can fetch tableware from the kitchen, and a legged robot can walk on firm pavement instead of unstable vegetation. However, even with semantics, maps can still be limited for higher level planning. Planning systems may desire other environmental attributes for better decision making, such as the traversability of the ground surface, the affordance~\cite[Ch. 8]{gibson2014ecological} of surrounding objects, the outdoor temperature, or the concentration of some quantity in the air. These attributes form a high-level understanding of the environment, which is regarded as one of the key requirements of the \emph{robust-perception age} in robotics~\cite{cadena2016past}.

At present, each of these attributes is often estimated independently and represented in a separate map. To perform a complex task involving multiple attributes, a robot needs to access and process information across different maps~\cite{verdoja2020potential}. This not only complicates planning, but ignores the correlation among different sources of information. In this paper, we argue that instead of using an independent map for a specific task, information can be stored in one shared map with several correlated layers where each layer contains one attribute. We propose a multitask multilayer Bayesian mapping framework which
\begin{enumerate}
    \item provides rich information such as semantics and traversability and is extendable to include more layers;
    \item runs online using an arbitrary number of sensors (e.g., cameras and LiDARs);
    \item unifies and simplifies the access to different map attributes for communication efficiency;
    \item leverages inter-layer correlations to improve the map posterior distribution.
\end{enumerate}
This work focuses on applying the proposed mapping framework to build a second \emph{traversability} layer on top of the first \emph{semantic} layer developed in our previous work~\cite{gan2020bayesian}. Traversability, by definition in robotics, means ``the capability of a ground robot to reside over a terrain region under an admissible state wherein it is capable of entering given its current state''~\cite{papadakis2013terrain}. We choose traversability as the second layer for two reasons. First, a traversability map is essential for a robot to safely and efficiently navigate within its environment and plan local motion policies with improved stability. Secondly, traversability is highly correlated with the semantic properties of an environment. For example, sky is non-traversable to a ground robot; road is the opposite; mud can either be traversable or untraversable.

An overview of the proposed framework is shown in Fig.~\ref{fig:overview}. We first employ exteroceptive sensor data and robot trajectory to generate ground truth labels for image traversability segmentation in a self-supervised manner. We then design and train a multitask deep neural network as our front-end to provide multiple observations simultaneously. Our back-end runs a closed-form Bayesian inference for dense multilayer mapping where semantic information is leveraged in traversability estimation. The multitask learning (MTL) front-end not only achieves systematic efficiency, but also improves prediction performance over the Single-Task Learning (STL) equivalent. The multilayer Bayesian inference back-end is scalable and provides map uncertainty.

This work has the following contributions:
\begin{enumerate}
    \item We propose a novel and extendable multitask multilayer Bayesian mapping framework to provide rich environmental information in a single mapping formalism.
    \item We develop a multitask deep neural network for scene semantic and traversability segmentation.
    \item We present a self-supervised approach to automatically generating training data for traversability segmentation.
    \item We formulate a scalable and dense multilayer mapping algorithm via closed-form Bayesian inference that leverages inter-layer correlations.
    \item We provide an open-source implementation of the proposed method and present extensive experiments using real and simulated data. We also discuss the reproducibility and extendability of the developed mapping framework.
\end{enumerate}

The remainder of this paper is organized as follows. A review of the related work on multilayer robotic mapping, self-supervised learning for traversability estimation and deep MTL is given in \secref{sec:literature}. \secref{sec:method} describes the design of our multitask deep network, and how we generate labeled traversability data for training without requiring manual annotation. \secref{sec:multi_layer_mapping} formulates our multilayer Bayesian map inference in the semantic-traversability case. Experiments and discussions are presented in \secref{sec:evaluation} and \secref{sec:discussion}, and finally, \secref{sec:conclusion} concludes the paper.

\section{Related Work}
\label{sec:literature}
In this section, we review the existing literature on multilayer robotic mapping, self-supervised traversability estimation, and deep MTL.

\subsection{Multilayer Robotic Mapping}

Multilayer robotic mapping, also known as hybrid mapping, stems from the work of~\citet{kuipers1991robot} where a hierarchical map with a global (middle-level) \emph{topological} layer and a local (low-level) \emph{metric} layer is built in robot exploration. This layout has been widely used in early works of multilayer mapping~\cite{thrun1998learning, bosse2003atlas, tomatis2003hybrid}. These works rarely contain multiple attributes observed by different sensory modalities, but rather abstract topological relationships from geometric information (i.e., homogeneous). Later, a high-level \emph{conceptual} layer is added to form the \emph{spatial semantic hierarchy}~\cite{kuipers2000spatial, galindo2005multi, zender2008conceptual}, where semantic knowledge extracted from vision or dialogue is integrated. Therefrom, multilayer robotic maps include environmental attributes more than geometry (i.e., heterogeneous), and are built from multiple sensory modalities.

More recently, computer vision and machine learning advancements significantly increase the potential of multilayer robotic mapping. \citet{pronobis2012large} develop a probabilistic framework based on chain graphs to build a four-layer map in three hierarchy levels using multi-modal sensory data: the low-level sensory layer, the middle-level place layer and categorical layer, and the high-level conceptual layer. \citet{jiang2019flexible} propose a four-layer lane-level map model for autonomous vehicles, where each layer contains different types of data and is dedicated to different navigation tasks. For instance, the first road layer is used for static mission planning and the fourth lane layer for reference trajectory planning. A multilayer High-Definition (HD) map consisting of a road graph, lane geometry and semantic features is also the convention in today's self-driving industry.

Scene graph, a data structure commonly used for describing 3D environments in computer graphics, has been employed in robotics as another form of multilayer map. In a scene graph, each layer has a set of \emph{nodes} representing entities and \emph{edges} between nodes indicating entity relations; each node is also associated to some \emph{attributes}. \citet{armeni20193d} construct a four-layer graph that spans an entire building and includes semantics on objects, rooms and cameras, as well as the relationships among them.
\citet{rosinol20203d} further propose 3D dynamic scene graphs to handle dynamic entities and capture actionable information in the environment. Early works in this direction pose challenges to general robotic navigation and exploration as scene graphs are usually constructed offline and the actionability is at an abstractive topological level. However, recent works~\cite{zheng2019pixels, rosinol2021kimera} have shown promising results for online scene graph construction that is readily usable for path planning. The work of~\cite{hughes2022hydra} is also a recent attempt at real-time 3D scene graph construction for indoor environments. The proposed work aims to improve the metric-semantic aspect of a more general robotic dense map.

DenseSLAM by \citet{nieto2006denseslam} is similar to our work as it also builds a dense grid-based multilayer map of the environment using a probabilistic framework. The fundamental difference is that it contains only geometric information acquired by range-finder sensors. The work by~\citet{nordin2011multi} is also related to ours in which a multilayer map with a geometric traversability layer and a ground roughness layer is built for path planning (based on traversability) and velocity control (based on roughness) of ground vehicle systems. However, the map does not provide any semantic knowledge of the environment or explore inter-layer correlations. More recently, \citet{zaenker2020hypermap} present a hypermap framework for autonomous semantic exploration. The map includes a grid-based occupancy layer, a polygonal semantic layer, and an exploration layer indicating the areas have yet to be explored. Our work differs from it in that we do not need an extra hypermap interface to convert and unify the information in each layer. Instead, we use the same grid-based representation for each layer to achieve unification.

\subsection{Self-Supervised Learning for Traversability Estimation}

Traversability analysis/estimation of the environment is crucial for autonomous navigation in a non-end-to-end framework. In the survey done by~\citet{papadakis2013terrain}, the traditional approaches to estimating traversability are categorized into \emph{geometry-based} analysis of digital elevation maps~\cite{wermelinger2016navigation} and \emph{appearance-based} classification of the terrain into a set of pre-defined classes using supervised learning~\cite{filitchkin2012feature}. However, supervised approaches are not robust to environmental changes and unsustainable for the widespread deployment of robots. They require a significant amount of manual labeling effort to adapt to different distributions. Self-supervised learning, in contrast, using a reliable module to generate supervisory signals for training another model by exploiting the correlations between different input signals, not only automates the labeling process but also gains flexibility in changing environments and platform variations (as the supervisory signals also vary accordingly under different conditions). We note that for some tasks of which the labels do not change much across environments and platforms, such as semantic segmentation, supervised learning approaches are still predominant.

\begin{figure}[t]
\centering
\includegraphics[width=0.99\columnwidth, trim={0.5cm 10cm 32cm 1.5cm}, clip]{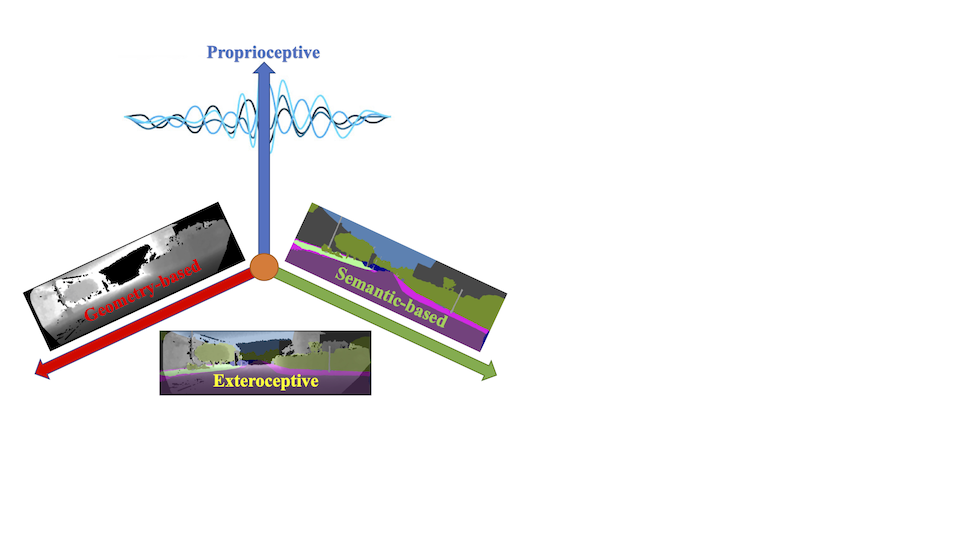}
    \caption{Space decomposition of self-supervised traversability learning methodologies based on the type of supervisory signals employed, i.e., proprioceptive, geometry-based and semantic-based approaches, and the latter two comprising the domain of exteroceptive approaches.}
    \label{fig:traversability_coordinate}
\end{figure}

Similar to~\cite{papadakis2013terrain}, we can project these methodologies onto a space characterized in Fig~\ref{fig:traversability_coordinate}, based on the type of supervisory signals employed in self-supervised traversability learning methods. We note that for traversability estimation problems, where we usually need to \emph{predict} the traversability of the ground areas have yet to be traversed, no matter what type of supervisory signals is employed, the \emph{deployed} sensory data tends to be exteroceptive.

Proprioceptive sensors that measure the internal states of the robot, e.g., acceleration~\cite{stavens2012self, bekhti2020regressed}, force~\cite{brooks2012self, wellhausen2019should}, torque~\cite{wellhausen2019should} and vibration~\cite{brooks2012self}, are commonly used to generate supervisory signals for traversability as they directly reflect the physical robot-terrain interaction. \citet{stavens2012self} generate the label of terrain roughness automatically from a vehicle's inertial measurements while driving. Acceleration signals are used as labels for traversability cost regression based on image textures in~\cite{bekhti2020regressed}. \citet{brooks2012self} train two proprioceptive classifiers based on a traction force model and wheel vibration. \citet{wellhausen2019should} record force-torque signals from foot-mounted sensors and images from an onboard camera equipped by a legged robot. Label values generated from the force-torque signals are then projected onto the images to train a convolutional network for terrain property prediction. Recently, robot-terrain interaction sounds captured by microphones are also used as a proprioceptive modality to self-supervise a visual terrain classification network~\cite{zurn2020self}.

A common drawback of using proprioceptive data as supervisory signals is that labels can only be generated for the areas directly interacted with the robot. Heuristics may be necessary for generating negative samples. Thus, exteroceptive sensor (e.g., camera, LiDAR) data is also used for self-supervision. Stanley won the 2005 DARPA Grand Challenge by employing a self-supervised road detection algorithm that uses laser range finders to identify and label drivable areas in the corresponding images~\cite{dahlkamp2006self}. Stereo images are used to self-supervise a vision-based traversability classifier in a near-to-far setting on a DARPA program LAGR platform~\cite{hadsell2009learning}. To train a deep segmentation network for path proposals in urban environments, \citet{barnes2017find} project vehicle future paths (positive) and obstacles detected by a LiDAR scanner (negative) onto images for labeling. Similarly, \citet{broome2020road} input geometric features extracted from accumulated LiDAR point clouds to a fuzzy-logic rule set to label the traversability for radar frames automatically. However, all these works only use geometric information acquired by the exteroceptive sensors. We use both geometric and semantic information extracted from exteroceptive sensory data for self-supervised traversability labeling. In addition, we also exploit MTL and multilayer mapping to improve the traversability estimation accuracy.

\subsection{Deep Multitask Learning}

MTL shares the same philosophy as our multilayer map inference, aiming to improve the performance of multiple related learning tasks by leveraging useful information among them~\cite{ruder2017overview}. Theoretically, MTL models induce a preference for hypotheses that explain multiple tasks, thus leading to reduced overfitting and improved generalization capabilities. Although MTL emerged long before the advent of deep learning~\cite{caruana1997multitask}, it has received more attention recently due to more promising advantages in combination with deep learning, e.g., improved data/memory efficiency and learning/inference speed.

In the deep learning era, MTL equates to designing Deep Neural Networks (DNNs) capable of learning shared representations from multitask signals~\cite{vandenhende2020multi}. Particularly, three problems have been mainly studied in this field: network architectures, optimization strategies, and task relationships~\cite{vandenhende2020multi, crawshaw2020multi}. For architectures, the key questions to answer are \emph{what to share} and \emph{how to share} parameters among tasks. Existing MTL methods are often categorized into \emph{hard} and \emph{soft parameter sharing} networks~\cite{ruder2017overview}. Hard parameter sharing networks share the exactly same model weights in shared layers~\cite{teichmann2018multinet}, while in soft parameter sharing networks, tasks have separate weights but interact with each other through regularization~\cite{yang2016trace}, cross-talk~\cite{misra2016cross, gao2019nddr}, or attention~\cite{liu2019end, maninis2019attentive} in recent works. A new taxonomy of MTL architectures is proposed in~\cite{vandenhende2020multi}, grouping them into \emph{encoder-focused} and \emph{decoder-focused} networks based on where the task interactions take place. It is noteworthy that MTL network architecture can also be learned automatically and dynamically from data instead of being manually designed and fixed~\cite{lu2017fully, rosenbaum2017routing}.

However, with an increase in the number and gap of tasks being studied, a well-known issue arises in MTL as \emph{negative transfer}. Negative transfer refers to the phenomenon of performance degradation when sharing information among unrelated or loosely-related tasks. Studies on optimization strategies and task relationships are the effort to tackle this problem. The main stream of works on optimization strategies treat MTL as a single-objective optimization problem with a weighted sum of task-specific losses, and focus on task balancing in terms of these weights. Uncertainty weighting~\cite{kendall2018multi} and Gradient Normalization (GradNorm)~\cite{chen2018gradnorm} are two popular approaches among them. Another stream of works formulate MTL as a multi-objective optimization problem and try to find a Pareto optimal solution among all tasks~\cite{sener2018multi}. On the other hand, task relationship learning also directs a way to address negative transfer by leveraging task relatedness to choose tasks to be learned together~\cite{standley2019tasks}, or to design network architectures. As this topic falls outside the scope of our work, we refer the interested reader to the more detailed discussion in~\cite{ruder2017overview}.

Attention mechanism has often been used in deep learning to visualize and interpret the inner states of convolutional neural networks, and been successfully applied to natural language processing tasks. More recently, the usage of attention in MTL achieves promising performances~\cite{liu2019end, maninis2019attentive}. As soft parameter sharing methods, attention-based networks use soft and differentiable attention masks to select/modulate features for each task from a shared backbone. They allow each task to use the shared representation differently, which alleviates negative transfer in hard parameter sharing networks where the shared features are constrained to be the same. As attention modules are small compared to the network backbone, these methods also do not suffer from the scalability issue in common soft parameter sharing networks where each task owns a completely separate set of parameters. Furthermore, attention modules can be incorporated into any feed-forward backbone architectures and learned in an end-to-end manner.

\section{Unified Segmentation Model for Multilayer Mapping via Multitask Learning}
\label{sec:method}

Robotic maps containing only geometric information usually directly take sensor (e.g., camera, LiDAR, radar, sonar) data as inputs to their mapping systems. However, with the emergence of semantic mapping, a reasoning block that can interpret the raw sensor data for a higher-level understanding of the scene has become essential to the mapping pipeline. This reasoning block is usually chosen as a deep neural network for place recognition, object detection, or semantic segmentation, depending on the specific mapping framework.

As most produced maps only contain one environmental attribute (e.g., room-level, object-level, or dense semantics), the employed DNN is trained for a single task. However, in a multilayer mapping system, it is inefficient or even unrealistic to have multiple DNNs, one per layer. Therefore, we propose to design a unified DNN model for different environmental/map attributes functioning as our reasoning block via MTL. This approach also justifies a practical and real-world application of MTL research in robotics.

In this section, a multitask DNN for image segmentation is presented. The MTL network takes monocular RGB images as inputs, and outputs several dense segmentation results to provide useful measurements for the following multilayer map inference. We first describe the general architecture design and model objective of the network. Then, we present a specific example of training the network for scene semantic and traversability segmentation. Semantic ground truth labels are publicly available in common data sets, whereas traversability labels are rare and usually robot-dependent. An automated traversability labeling approach using exteroceptive sensory data is also proposed in this section to tackle the issue. Based on how the labels are generated, we regard semantic segmentation as a supervised learning procedure while traversability segmentation self-supervised. Furthermore, the training procedures are generalizable to other map attributes, given the corresponding ground truth labels.

\subsection{Multitask Network with Attention Mechanisms}

\subsubsection{Architecture Design}

Inspired by recent works on MTL using attention~\cite{liu2019end, maninis2019attentive}, our network consists of two components: a single shared encoder with task-specific attention modules, and $K$ task-specific decoders for $K$ tasks. This design combines hard and soft parameter sharing strategies: each task has its individual decoder and an attention-modulated shared encoder. It reduces the risk of overfitting by encouraging the encoder to learn to extract features that can fit all tasks. In the meantime, it leverages attention mechanisms to alleviate the negative transfer issue when the encoder is shared among less related or even conflicting tasks.

We adopt two attention mechanisms following the work of~\citet{maninis2019attentive}: Squeeze-and-Excitation (SE) blocks and Residual Adapters (RA). Squeeze-and-excitation block is proposed by~\citet{hu2018squeeze} to adaptively recalibrate channel-wise feature responses by explicitly modeling interdependencies between channels. In other words, it is a lightweight gating mechanism in channel-wise relationships. SE block has a \emph{squeeze} operator (usually an average pooling per channel) which aggregates feature maps into a channel descriptor, followed by an \emph{excitation} operator (two fully-connected layers around non-linearities) that maps the descriptor to a set of channel weights. Finally, the original feature maps are reweighted with the produced channel weights.

The idea of residual adapters is proposed by~\citet{rebuffi2017learning} for multi-domain learning. A small number of domain-specific parameters are added to the adapters to tailor the network to diverse visual domains, while a high percentage of parameter sharing among domains is still maintained. As multidomain learning can be considered as a specific MTL problem, RAs have been applied to MTL for adapting and refining the shared features of each task~\cite{maninis2019attentive}. RA module is essentially a $1 \times 1$ filter bank in parallel with a skip connection, introducing a small amount of computational overhead while substantially improving accuracy and alleviating overfitting~\cite{rebuffi2017learning}. We choose a more efficient variant with parallel adapters~\cite{rebuffi2018efficient} in our network. We set the SE and RA parameters to be task-dependent to allow each task to learn its own channel-wise relationships, as well as utilize the shared encoder differently.

\begin{figure}[t]
\includegraphics[width=0.99\columnwidth, trim={5cm 0cm 3.7cm 1.5cm}, clip]{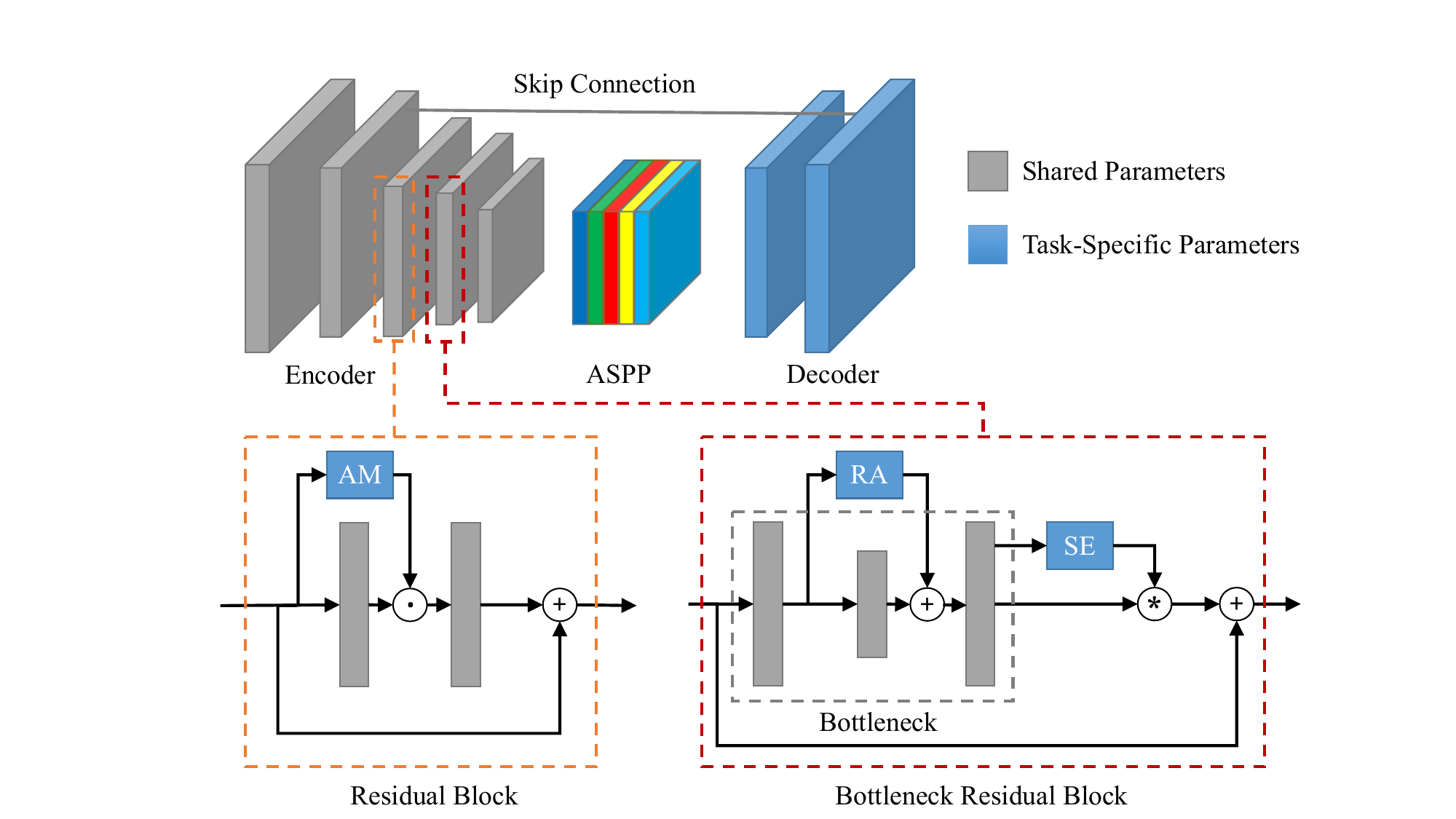}
\caption{Network architecture of each task using an illustrative backbone, where $\odot$, $+$ and $*$ denote operations of element-wise multiplication, addition, and matrix multiplication, respectively. The network has task-specific decoders and task-specific attention modules attached to the shared encoder. Attention modules consist of Residual Adapter (RA) and Squeeze-and-Excitation (SE) in the bottleneck residual blocks and Attention Mask (AM) in all other residual blocks. This attentive multitasking architecture can be built upon any feed-forward neural network with residual blocks and can be trained end-to-end.}
\label{fig:networks}
\end{figure}

Different from~\cite{maninis2019attentive}, we implement the task-specific SE and RA modules inside the \emph{bottleneck} blocks (one $3\times3$ convolution in the middle of two $1\times1$ convolutions) of a shared residual encoder, as shown in Fig.~\ref{fig:networks}. To better utilize the shared features extracted by other residual blocks, we add a task-specific element-wise Attention Mask (AM) to those blocks. The attention mask is chosen to be light-weighted, consisting of a $1\times1$ convolutional filter, a batch normalization layer, and a sigmoid activation to ensure the output belongs to~$[0, 1]$. We note that both RA and AM are implemented using convolutions, but the output of RA is added to, whereas the output of AM is element-wise multiplied by the shared features. The following equations represent the difference:
\begin{equation}
    F_k(x) = \text{RA}_k(x) + F(x),
\end{equation}
\begin{equation}
    F_k(x) = \text{AM}_k(x) \odot F(x),
\end{equation}
where $F(x)$ and $F_i(x)$ represent the original shared features and the modified task-specific features of task $k$, respectively, and $\odot$ denotes element-wise multiplication.

Fig.~\ref{fig:networks} illustrates the general architecture of each task in our network. Each task has its dedicated SE, RA and AM modules in the shared encoder to select and refine the shared parameters in a way more favorable to the specific task, with a relatively low parameter overhead. The modified features output from the entire shared encoder are fed into task-specific decoders. In our specific implementation, we have an Atrous Spatial Pyramid Pooling (ASPP)~\cite{chen2017deeplab} segmentation module between the shared encoder and task-specific decoders for feature resampling, but this MTL architecture can be built on any feed-forward neural network with residual blocks. All task-shared and task-specific parameters are learned in an end-to-end manner.

\begin{figure*}[t]
\centering
\includegraphics[width=0.9\textwidth, trim={0cm 8cm 0cm 2cm}, clip]{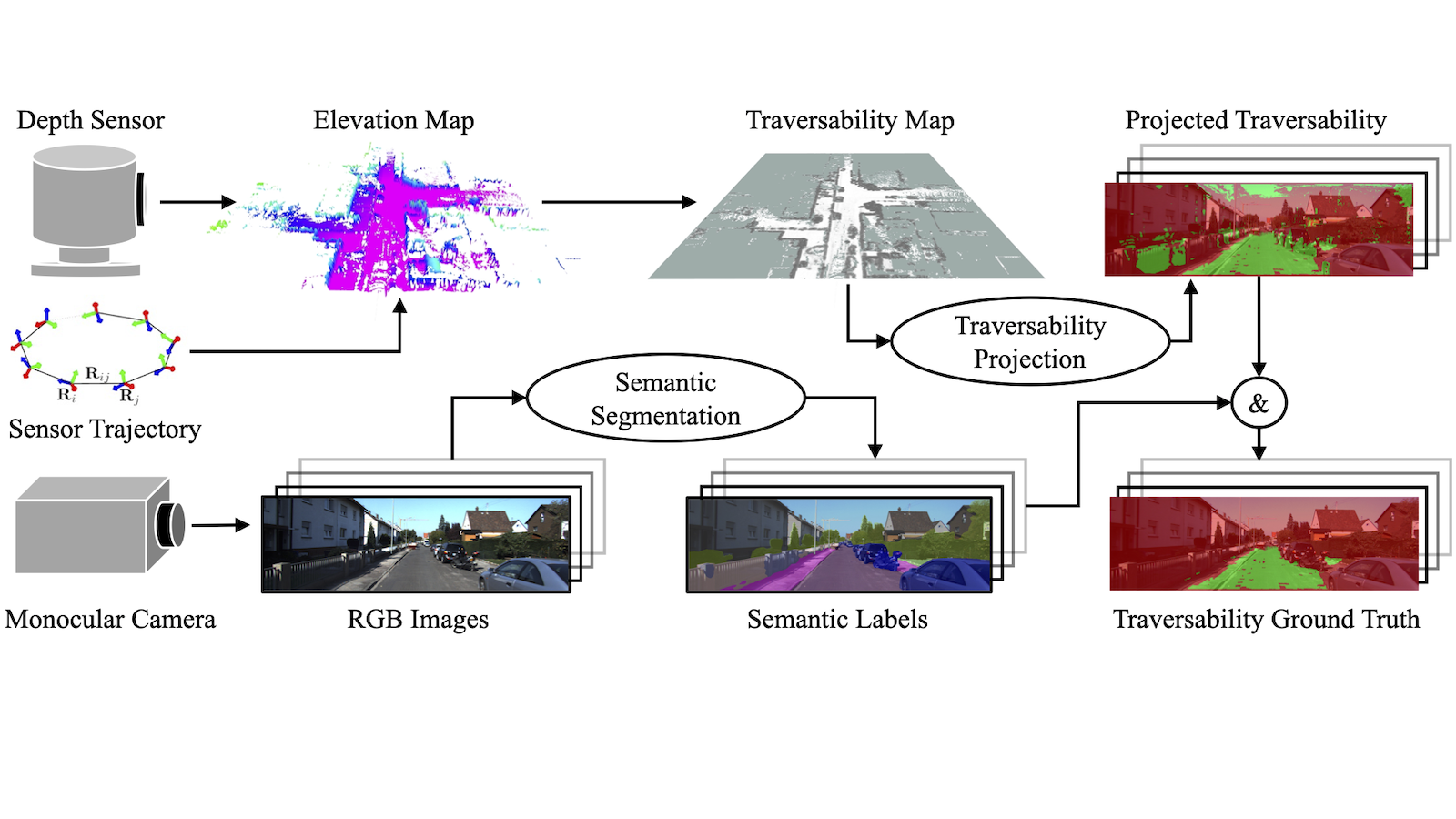}
    \caption{Our self-supervised traversability labeling pipeline in which exteroceptive sensory data is used as supervisory signals. The upper part concerns geometry; An elevation map is built using point cloud data captured by a depth sensor (LiDAR, depth camera or stereo camera) and the sensor trajectory. A 2D traversability map is then computed from the elevation map and is projected onto images. The lower part considers semantics; We use the predicted semantic labels to filter out (false positive) noises in the projected traversability images. Combining both geometry-based and semantic-based self-supervised approaches, the final traversability ground truth labels can be generated automatically and accurately (green is for traversable area and red for untraversable area).}
    \label{fig:trav_labeling}
\end{figure*}

\subsubsection{Model Objective}

In an MTL setting, the optimization objective for $K$ tasks is usually formulated as a weighted sum of all task-specific losses:
\begin{equation}
    \mathcal{L}_{\texttt{MTL}}(\mathbf{X}, \mathbf{Y}_{1:K}) = \sum_{k=1}^{K} w_k \mathcal{L}_k (\mathbf{X}, \mathbf{Y}_k),
\label{eq:total_loss}
\end{equation}
where $\mathbf{X}$ is the input, $\mathbf{Y}_k$ and $w_k$ are the ground truth label and task-specific loss weight of task $k$,~\mbox{$k = 1,2,...,K$}, respectively. For \emph{heterogeneous} MTL, where tasks include classification and regression problems, the task-specific loss $\mathcal{L}_k$ has different formulation according to the type of problem. As we are considering a \emph{homogeneous} MTL problem that consists of image segmentation tasks (i.e., classification), the task-specific loss function is chosen to be class-weighted one-hot label cross-entropy loss:
\begin{equation}
    \mathcal{L}_{k}(\mathbf{X}, \mathbf{Y}_{k}) = - \sum_{c=1}^{C_k} w_{k,c} \mathbf{Y}_{k,c} \log \hat{\mathbf{Y}}_{k,c},
\end{equation}
where $\mathbf{Y}_{k,c}$ is a binary variable indicating whether the ground truth label $\mathbf{Y}_k$ is class $c$, and $\hat{\mathbf{Y}}_{k,c}$ is the network predicted likelihood of $\mathbf{X}$ belonging to class $c$,~\mbox{$c=1,2,...,C_k$}. In order to compensate for unbalanced training data, we also weight the loss according to the number of samples in each class~\cite{long2015fully}, i.e., class weight $\lambda_{k,c}$ will be higher if class $c$ has fewer samples. Precisely, $\lambda_{k,c}$ is set to be the reciprocal of the frequency of class $c$ in training data, and is not part of the network parameters.

\subsection{MTL for Semantic and Traversability Segmentation}

The previous subsection proposed a generic MTL network with attention mechanisms that can be used as a unified segmentation model in our multilayer mapping system. This subsection describes a specific implementation of the network for scene semantic and traversability segmentation to provide the following mapping module with necessary measurements.

For segmentation tasks, we choose DeepLabv3+ architecture~\cite{chen2018encoder} with WideResNet38 backbone~\cite{wu2019wider} as this combination is used in~\cite{zhu2019improving} and achieves the state-of-the-art semantic segmentation performance on KITTI dataset~\cite{geiger2012we}. In our MTL setting, the semantic/traversability segmentation task has its individual decoder and its own set of attention modules attached to the shared encoder (the shared layers and task-specific layers are shown in different colors in Fig.~\ref{fig:networks}). The design of task-specific modules is the same for all tasks, except that the number of output channels of the last convolutional layer is set to be the number of classes in that task. We consider the traversability segmentation task as a pixel-wise binary classification problem (i.e., traversable vs. untraversable); thus, the number of classes in this task is two.

During training, all network parameters (shared and task-specific) are trained together using input images, semantic and traversability ground truth labels simultaneously. The following subsection describes an approach to automatically generating traversability ground truth labels used for training and evaluation in our experiments.

\subsection{Self-Supervised Traversability Labeling}
\label{sec:trav_labeling}

This subsection presents a self-supervised approach to generating vast quantities of labeled traversability training data using exteroceptive sensing without manual annotation. Specifically, our method uses image data, 3D sensing data acquired by the robot platform, and robot trajectories as supervisory signals for traversability. The main idea is to obtain the traversability labels of the environment from 3D geometry and transfer them to the 2D image domain for time-coherent annotation. The overview of our self-supervised traversability labeling process is illustrated in Fig.~\ref{fig:trav_labeling}. We first determine the pixel-wise traversability labels of an image based on its corresponding 3D geometry and robot capabilities, and then use available semantic information to filter out the label noises.

\subsubsection{3D Geometry-based Traversability Computation} 

We employ a geometry-based traversability estimation method with low complexity proposed by \citet{wermelinger2016navigation}. The traversability is estimated based on terrain characteristics and the traversing capabilities of the robot. A robot-centric \emph{elevation map} of the terrain is built from 3D measurements and 6-DoF poses of an exteroceptive sensor (such as a laser scanner, a depth camera, or a stereo camera). The elevation map is a 2.5D grid map in which each cell stores a height estimation and its variance~\cite{fankhauser2018probabilistic}.

Three typical local terrain characteristics, i.e., the \emph{slope} $s$, \emph{roughness} $r$, and \emph{step height} $h$, are computed for each map cell by applying different filters to the elevation map. The local traversability score~\mbox{$t_{\texttt{score}} \in [0, 1]$} of that cell, with higher value meaning more traversable, combines these characteristics as follows:
\begin{equation}
    t_{\texttt{score}} = 1 - w_1 \frac{s}{s_{\texttt{crit}}} - w_2 \frac{r}{r_{\texttt{crit}}} - w_3 \frac{h}{h_{\texttt{crit}}},
\label{eq:traversability}
\end{equation}
where $w_1$, $w_2$ and $w_3$ are the weights that add up to 1. The $s_{\texttt{crit}}$, $r_{\texttt{crit}}$ and $h_{\texttt{crit}}$ are the robot-specific maximum allowed value for these characteristics. The traversability score is set to be $0$ if one of $s$, $r$ and $h$ exceeds its critical value.

To generate dense labels for an image at timestamp $t$, we first build a local elevation map using successive frames of point cloud data from timestamp $t$ to \mbox{$t + l$} and the corresponding sensor poses, where $l$ is the time window length chosen based on data acquisition rate. Essentially, we build a local elevation map that aligns with the camera view at timestamp~$t$. We then compute the traversability score of each cell using \eqref{eq:traversability}, and result in a 2D local traversability map at timestamp $t$ that has the same size and resolution as the input elevation map.

\subsubsection{Map-to-Image Projection}

To project the traversability map onto the image plane, we use each pixel's depth value to retrieve the corresponding traversability score from the map and associate it with that pixel. In this way, we generate an image where each pixel has a traversability score, except those with invalid depth values. We then set a threshold for the projected score to get a binary label. An example of the projected traversability labels is given in Fig.~\ref{fig:trav_labeling}. 

\subsubsection{Semantics-based Noise Filtering}

The traversability labels generated so far are purely from geometric considerations, which can be quite noisy due to errors in sensor reading or depth/pose estimation. As shown in Fig.~\ref{fig:trav_labeling}, the projected traversability mislabels some sky pixels as traversable. To correct these false positive labels, we use semantic segmentation of the input image to filter out noises present in semantically untraversable areas, such as sky, buildings, and humans. After semantics-based noise filtering, the final traversability labels in Fig.~\ref{fig:trav_labeling} do not contain these false positives.

\section{Multilayer Bayesian Map Inference}
\label{sec:multi_layer_mapping}

Multilayer mapping aims to use the semantic occupancy map as \emph{a priori} knowledge or complementary information in modeling other environmental attributes, such as traversability, friction, and concentration of a physical quantity, by considering their correlations with occupancy and semantics. To this end, we extend our previous work for semantic mapping~\cite{gan2020bayesian} to a closed-form Bayesian inference method for multilayer mapping. This section first briefly introduces the semantic layer construction, and then describes how it can be leveraged in traversability layer inference. Extension of the method to more map layers is discussed in Section~\ref{sec:discussion_extensions}.

\subsection{Continuous Semantic Mapping}
\label{sec:semantic_mapping}

Assuming the 3D map cells are indexed by~\mbox{$j \in \mathbb{Z}^+$}, the $j$-th map cell with semantic probability $\theta_j$ can be described by a Categorical distribution (i.e., the \emph{likelihood}):
\begin{equation}
    p(y_i | \theta_j) = \prod_{k=1}^{K} \left ( {\theta_j^k} \right )^{y_i^k},
\label{eq:categorical}
\end{equation}
where~\mbox{$\theta_j = (\theta_j^1, ..., \theta_j^K)$, $\sum_{k=1}^{K} \theta_j^k = 1$} denote the probability of the $j$-th map cell taking the $k$-th category from a set of semantic categories~\mbox{$\mathcal{K} = \{1, 2, ..., K\}$}. The one-hot encoded semantic measurement~\mbox{$y_i = (y_i^1, ..., y_i^K)$} is from the semantic segmentation result of the MTL network.

Let \mbox{$\mathcal{X} \subset \mathbb{R}^3$} be the map spatial support, i.e., the Cartesian coordinates in 3D Euclidean space, the training set (data) for semantic map inference can be defined as \mbox{$\mathcal{D}_y := \{(x_i, y_i)\}_{i=1}^N$}, where \mbox{$x_i \in \mathcal{X}$} is the position of the $i$-th measurement and $N$ is the number of training points. In Bayesian semantic mapping, we seek the \emph{posterior} over $\theta_j$; \mbox{$p(\theta_j | \mathcal{D}_y)$}.

For a closed-form solution, we place a Dirichlet distribution over $\theta_j$ as the conjugate \emph{prior} of the Categorical likelihood, denoted by~\mbox{$\text{Dir}(K, \alpha_0)$}, where~\mbox{$\alpha_0 = (\alpha_0^1, ..., \alpha_0^K)$, $\alpha_0^k \in \mathbb{R}^{+}$} are the concentration parameters. By applying Bayes' rule and Bayesian kernel inference~\cite{vega2014nonparametric}, the posterior~\mbox{$\text{Dir}(K, \alpha_\ast)$}, $\alpha_\ast = (\alpha_\ast^1, ..., \alpha_\ast^K)$ can be computed as follows:
\begin{equation}
\label{eq:alpha_kernel}
    \alpha_\ast^k = \alpha_0^k + \sum_{i=1}^N k(x_\ast, x_i) y_i^k ,
\end{equation}
where \mbox{$k(\cdot, \cdot): \mathcal{X} \times \mathcal{X} \to [0, 1]$} is a kernel function operating on the map query point $x_\ast$ and the $i$-th training point $x_i$. To exactly and efficiently evaluate the kernel over relevant measurements (i.e., the measurements falling into the neighborhood of a query point), a sparse kernel~\cite{melkumyan2009sparse} is employed with the $k$-d tree data structure to achieve map continuity and $\mathcal{O}(\log N)$ computation time for a single map query, where $N$ is the number of training points.
For details, interested readers can refer to~\cite[and references therein]{gan2020bayesian}. 

\subsection{Semantic-Traversability Mapping}
\label{sec:s_t_mapping}

Bernoulli distribution is a natural and common choice for modeling binary variables such as occupancy~\cite{doherty2019learning}, and terrain traversability~\cite{shan2018bayesian}. For the $j$-th map cell, we treat its traversability as a Bernoulli distributed random variable which takes the value 1 (i.e., traversable) with probability $\phi_j$. The measurement likelihood is described as:
\begin{equation}
\label{eq:traversability_likelihood}
    p(z_i | \phi_j ) = \phi_j^{z_i}(1-\phi_j)^{1-z_i} ,
\end{equation}
where the binary measurements~\mbox{$\mathcal{Z}=\{ z_1, ..., z_N | z_i \in \{0, 1\}\}$} indicate whether the map cell containing the corresponding position in~\mbox{$\mathcal{X} = \{ x_1, ..., x_N | x_i \in \mathbb{R}^3 \}$} is classified as untraversable or traversable by the neural network. Given training set \mbox{$\mathcal{D}_z := \{(x_i, z_i)\}_{i=1}^N$}, Bayesian traversability mapping is thus formulated as seeking the posterior distribution $p(\phi_j | \mathcal{D}_z)$ of each map cell.

\begin{algorithm}[t]
\caption{Semantic Traversability Bayesian Inference}
\label{al:s_t_inference}
\begin{algorithmic}[1]
\State \textbf{Input:} Training data: $\mathcal{X}, \mathcal{Y}, \mathcal{Z}$; Query point: $x_\ast$
\State \textbf{Initialize:} semantic prior parameters $\alpha_\ast^k \gets 0.001$
\State \hspace{1.45cm} traversability prior parameters $\alpha_\ast, \beta_\ast \gets 0.001$ 
\For{each $(x_i, y_i) \in (\mathcal{X}, \mathcal{Y})$}
    \State $k_i \gets k(x_\ast, x_i)$
    \For{$k=1,...,K$}
        \State $\alpha_\ast^k \gets \alpha_\ast^k + k_i y_i^k$
    \EndFor
\EndFor \Comment{Semantic mapping}
\For{each $(x_i, z_i) \in (\mathcal{X}, \mathcal{Z})$}
    \If{$x_i$ in cell $j$}
        \State $\alpha_j^\prime, \beta_j^\prime \gets 0$
        \For{$k=1,...,K$}
            \If{$k$-th category is traversable}
                \State $\alpha_j^\prime \gets \alpha_j^\prime + \alpha_j^k$
            \ElsIf{$k$-th category is untraversable}
                \State $\beta_j^\prime \gets \beta_j^\prime + \alpha_j^k$
            \EndIf
        \EndFor \Comment{Convert Dirichlet to beta distribution}
        \State $\phi_j^\prime \gets (\alpha_j^\prime - 1) / (\alpha_j^\prime + \beta_j^\prime -2)$  \Comment{Compute mode}
        \State $u \sim \mathcal{U}(0, 1)$ \Comment{Draw a uniformly distributed random number}
        \If{$u \leq \phi_j^\prime$}
            \State $y_i^\prime \gets 1$
        \Else{}
            \State $y_i^\prime \gets 0$
        \EndIf \Comment{Sample from Bernoulli distribution}
    \EndIf \Comment{Generate semantic-traversability measurements}
    \State $k_i \gets k(x_\ast, x_i)$
    \State $\alpha_\ast \gets \alpha_\ast + k_i (y_i^\prime + z_i)$
    \State $\beta_\ast \gets \beta_\ast + k_i (2 - y_i^\prime - z_i)$
\EndFor \Comment{Semantic-traversability mapping}
\State \textbf{return} $\alpha_\ast^k, \alpha_\ast, \beta_\ast$
\end{algorithmic}
\end{algorithm}

Semantics of the environment can provide an important insight into its traversability. For instance, we know the sky and buildings are untraversable to ground robots, while the road is traversable to an autonomous vehicle. Therefore, we argue that semantics can be used as another source of observation available for traversability mapping, and a better traversability inference performance is expected when the correlations are leveraged.

In semantic-traversability mapping, we first shrink the Dirichlet posterior distribution~\mbox{$\text{Dir}(K, \alpha_j)$, $\alpha_j = (\alpha_j^1, ..., \alpha_j^K)$} obtained from continuous semantic mapping to a beta distribution $\text{Beta}(\alpha_j^\prime, \beta_j^\prime)$, where $\alpha_j^\prime$ is the sum of concentration parameters $\alpha_j^k$ for all traversable categories $k$, and $\beta_j^\prime$ is the sum of $\alpha_j^k$ for all untraversable categories. As Dirichlet distribution is a multivariate generalization of the beta distribution, $\text{Beta}(\alpha_j^\prime, \beta_j^\prime)$ can be proved to be a valid beta distribution. 

From the beta distribution, we can deduce a Bernoulli distribution $\textnormal{Bernoulli}(\phi_j^\prime)$ to describe the traversability uncovered by the semantics, where \mbox{$\phi_j^\prime = (\alpha_j^\prime - 1) / (\alpha_j^\prime + \beta_j^\prime - 2)$}, \mbox{$\alpha_j^\prime, \beta_j^\prime > 1$} is a maximum a posteriori (MAP) estimate of $\phi_j^\prime$. We then sample from this Bernoulli distribution to generate semantic-traversability measurements \mbox{$\mathcal{Y}^\prime = \{y_1^\prime, ..., y_N^\prime | y_i^\prime \in \{0, 1\}\}$}. Specifically, for each training point $x_i$, we first find the corresponding map cell $j$ that contains $x_i$, and then generate a binary measurement $y_i^\prime$ according to the deduced $\textnormal{Bernoulli}(\phi_j^\prime)$ of that cell. We refer to the generated $\mathcal{Y}^\prime$ as the semantic-traversability measurements because they are indirect observations of traversability through semantics.

To incorporate both measurements into traversability inference, we assume that measurements $\mathcal{Y}^\prime$ also have the same likelihood as in~\eqref{eq:traversability_likelihood}:
\begin{equation}
    p(y^\prime_j | \phi_j ) = \phi_j^{y^\prime_j}(1-\phi_j)^{1-y^\prime_j}.
\end{equation}
We are now concerned with the posterior over possible $\phi_j$; $p(\phi_j | \mathcal{D}_{y^\prime}, \mathcal{D}_z)$. Assuming $\mathcal{D}_{y^\prime}$ and $\mathcal{D}_z$ are independent, we have:
\begin{equation}
    p(\phi_j | \mathcal{D}_{y^\prime}, \mathcal{D}_z) \propto 
    p( \phi_j | \mathcal{D}_{y^\prime}) p(\phi_j | \mathcal{D}_z ),
\end{equation}
where $\mathcal{D}_{y^\prime}$ comes from semantic segmentation and mapping, while $\mathcal{D}_z$ is directly obtained by traversability segmentation.

For a closed-form formulation, we adopt a conjugate prior of the Bernoulli likelihood as \mbox{$\phi_j \sim \textnormal{Beta}(\alpha_0, \beta_0)$}, in which \mbox{$\alpha_0 > 0$} and \mbox{$\beta_0 > 0$} are the shape parameters. According to Bayes' rule, we have \mbox{$\phi_j | \mathcal{D}_{y^\prime} \sim \textnormal{Beta}(\alpha_{y^\prime}, \beta_{y^\prime})$}, where $\alpha_{y^\prime}$ and $\beta_{y^\prime}$ are defined as follows:
\begin{align}
\alpha_{y^\prime} &:= \alpha_0 + \sum_{i, \text{ $x_i$ in cell } j} y_i^\prime \\
\beta_{y^\prime} &:= \beta_0 + \sum_{i, \text{ $x_i$ in cell } j} (1-y_i^\prime).
\end{align}
Then we have \mbox{$\phi_j | \mathcal{D}_{y^\prime}, \mathcal{D}_z \sim \textnormal{Beta}(\alpha_j, \beta_j)$}, where:
\begin{align}
\label{eq:alpha}
    \alpha_j &:= \alpha_{y^\prime} + \sum_{i, \text{ $x_i$ in cell } j} z_i\\
    \beta_j &:= \beta_{y^\prime} + \sum_{i, \text{ $x_i$ in cell } j} (1-z_i).
\end{align}

Applying Bayesian kernel inference~\cite{vega2014nonparametric}, the final traversability posterior $p(\phi_j | \mathcal{D}_{y^\prime}, \mathcal{D}_z)$ can be obtained as $\textnormal{Beta}(\alpha_j, \beta_j)$ with $\alpha_j$ and $\beta_j$ defined as:
\begin{align}
    \alpha_j &:= \alpha_0 + \sum_{i=1}^N k(x_j, x_i) (y^\prime_i + z_i) \\
    \beta_j &:= \beta_0 + \sum_{i=1}^N k(x_j, x_i) (2 - y^\prime_i - z_i).
\end{align}
The MAP estimate of $\phi_j$ then has the closed-form solution:
\begin{equation}
\label{eq:mode}
    \hat{\phi}_j = \frac{\alpha_j - 1}{\alpha_j + \beta_j - 2}\ \text{and}\ \alpha_j, \beta_j > 1.
\end{equation}
The expected value and variance of $\phi_j$ can also be computed in closed form:
\begin{equation}
\label{eq:variance}
    \mathbb{E}[\phi_j] = \frac{\alpha_j}{\alpha_j + \beta_j}\ \text{and}\
    \mathbb{V}[\phi_j] = \frac{\alpha_j \beta_j}{(\alpha_j + \beta_j)^2 (\alpha_j + \beta_j + 1)} .
\end{equation}

An algorithmic implementation of the semantic traversability Bayesian inference procedure is provided in Algorithm~\ref{al:s_t_inference}. Lines 4-9 are the inference procedure for semantic mapping and lines 10-31 are for semantic-traversability mapping. We use a simple sampling strategy (lines 21-26) to generate semantic-traversability (pseudo-)measurements from the Bernoulli distribution deduced from semantic posterior. In this way, we are able to leverage more accurate and up-to-date semantic posteriors in traversability inference, instead of the noisy single-frame semantic measurements.

The computational complexity for building the semantic layer is $\mathcal{O}(M\log N)$, where $M$ and $N$ is the number of test points and training points, respectively. The semantic-traversability mapping retains the same $\mathcal{O}(M\log N)$ time complexity as the semantic mapping, with small additional memory expenditures to store the deduced Bernoulli distributions and traversability posteriors. However, as we use the posterior distribution of one layer to build another layer, the computational complexity of the multilayer mapping grows linearly with the number of map layers.

\begin{remark}
 The traversability layer inference is similar to the continuous occupancy mapping approach of~\cite{doherty2019learning}. While the mathematical derivations are the same, the traversability layer in this work performs two updates per iteration; one essentially from the inferred semantic layer and one directly from a neural network.
\end{remark}

\begin{table*}
\centering
\caption{Ablation study of deep MTL networks for semantic and traversability segmentation. Mean intersection over unions (mIoUs) of 19 semantic classes and 2 traversability classes are reported on a test set from KITTI odometry sequence 10. Multitask performance $\Delta_\texttt{MTL}$ indicates the average per-task performance gain w.r.t the single-task baseline (see~\eqref{eq:mtl_metric}).}
\label{table:ablation_multitask}
\resizebox{0.95\textwidth}{!}{
\begin{tabular}{c l l c c c}
\toprule
\multirow{2}{*}{Setting} & \multirow{2}{*}{Model} & \multirow{2}{*}{Training} & \multicolumn{2}{c}{Segmentation mIoU ($\%$)} & \multirow{2}{*}{$\Delta_\texttt{MTL}$ ($\%$)}\\
& & & Semantic & Traversability & \\
\midrule
1 & STL & Individually for each task & 83.08 & 79.80 & + 0.00\\
2 & MTL w/o Attentions, Fixed & Simultaneously with shared parameters fixed & 83.08 & 77.06 & - 1.72 \\
3 & MTL w/o Attentions & Simultaneously & 83.24 & 78.38 & - 0.79 \\
4 & MTL w/ Attentions & Simultaneously &\bf 86.51 & \bf 83.91 & \bf + 4.64 \\
\bottomrule
\end{tabular}
}
\end{table*}

\section{Experimental Evaluation}
\label{sec:evaluation}

In this section, we evaluate our proposed method using two publicly available datasets for robotic applications: the KITTI dataset~\cite{geiger2012we} and the TartanAir dataset~\cite{wang2020tartanair}. For both datasets, qualitative and quantitative results are provided with discussion. At the end of this section, we further validate the proposed method using data collected by our bipedal robot platform, Cassie Blue.

\subsection{Implementation Details}
\label{sec:implementation}

For the MTL network, we use the architecture shown in Fig.~\ref{fig:networks} based on DeepLabv3+ and WideResNet38 backbone. During training, we use an SGD optimizer with initial learning rate 0.001, momentum 0.9 and weight decay 0.0001. A polynomial learning rate policy is also employed with the power of 1.0. In addition, we make the Synchronized Batch Normalization (SBN) in~\cite{zhu2019improving} task-specific, with a batch size of 8 distributed over two NVIDIA TITAN RTX GPUs. We use the pretrained semantic segmentation weights on Cityscapes~\cite{cordts2016cityscapes} for model initialization in all experiments. The traversability-specific decoder is also initialized with the pretrained weights considering the two tasks are closely related. All task-specific attention modules are randomly initialized. Our network implementation is based on~\cite{zhu2019improving} using PyTorch. We provide our code at~\href{https://github.com/ganlumomo/mtl-segmentation}{https://github.com/ganlumomo/mtl-segmentation}.

The multilayer Bayesian mapping algorithm is implemented in C++ with the Robot Operating System (ROS) and uses the Learning-Aided 3D Mapping (LA3DM) library~\cite{doherty2019learning}. Kernel length-scale and kernel scale of the sparse kernel are set to 0.3 $\m$ and 10, respectively, and remain fixed throughout the experiments. Dirichlet concentration parameters and beta shape parameters are initialized to 0.001. All mapping experiments are conducted on an Intel i7 processor with 8 cores and 32 GB RAM. The implementation of our multilayer mapping algorithm is available at~\href{https://github.com/ganlumomo/MultiLayerMapping}{https://github.com/ganlumomo/MultiLayerMapping}.

For self-supervised traversability labeling, we use the robot-centric elevation mapping library~\cite{fankhauser2018probabilistic} and the traversability estimation package~\cite{wermelinger2016navigation}. Dataset loaders for KITTI and TartanAir are provided, along with the code for projecting traversability scores onto the image plane. Elevation maps are built at a resolution of 0.1 $\m$. The $s_\texttt{crit}$, $r_\texttt{crit}$ and $h_\texttt{crit}$ in~\eqref{eq:traversability} are set to 1.0, 0.05 and 0.12, respectively. The weights are set to $1/3$, and the threshold for $t_\texttt{score}$ is 0.5. The whole labeling pipeline is implemented in ROS, and available at~\href{https://github.com/ganlumomo/traversability_labeling_ws/}{https://github.com/ganlumomo/traversability\_labeling\_ws}.

\subsection{Ablation Study}
\label{sec:ablation}

To better understand the effect of the task-specific attention modules, we first perform an ablation study of our MTL network with and without attention modules and compare the performance with the corresponding single-task learning baseline. For a fair comparison, we use the same architecture (DeepLabv3+ with WideResNet38 backbone) and pretrained model (on Cityscapes) for the following settings:

\begin{enumerate}
    \item \emph{STL}: The vanilla DeepLabv3+ architecture with WideResNet38 backbone for single-task learning. This model is proposed by~\citet{zhu2019improving} for semantic segmentation, and we also train it for traversability segmentation.
    \item \emph{MTL w/o Attentions, Fixed}: The standard MTL baseline with a shared encoder and task-specific decoders~\cite{vandenhende2020multi}, trained with the shared parameters fixed.
    \item \emph{MTL w/o Attentions}: The same architecture as above~\cite{vandenhende2020multi} with all parameters trained.
    \item \emph{MTL w/ Attentions}: The same architecture as above with additional task-specific attention modules and task-specific SBN operators attached to the shared encoder. All parameters are trained.
\end{enumerate}

As for evaluation criterion, in addition to the standard mean Intersection over Union (mIoU) for segmentation accuracy, we also use a multitask performance metric defined in~\cite{vandenhende2020multi}:
\begin{equation}
\label{eq:mtl_metric}
    \Delta_\texttt{MTL} = \frac{1}{K}\sum_{i=1}^{K} (-1)^{l_i} (M_{m,i} - M_{s,i}) / M_{s,i},
\end{equation}
where $M_{m,i}$ and $M_{s, i}$ are the performance of the MTL and STL model on task $i$, respectively, and $K$ is the number of tasks. $l_i = 0$ if a higher value means a better performance for metric $M_i$, and 1 otherwise. A higher $\Delta_\texttt{MTL}$ means a better multitask performance w.r.t the same STL baseline.

We train these four models using the same hyperparameters on KITTI dataset~\cite{geiger2012we}. For semantic segmentation, we use the KITTI pixel-level semantic dataset consisting of 200 annotated training images. We randomly select 20 images for testing and exclude them from the training process. For traversability segmentation, we automatically generate traversability labels for 200 images from sequence 06 and 07 of the KITTI odometry dataset for training, and another 200 images from sequence 10 for testing. The ablation study results are reported in Table~\ref{table:ablation_multitask}.

From the results in Table~\ref{table:ablation_multitask}, the standard MTL baselines (setting 2 and 3) fail to outperform their single-task equivalent (setting 1) for traversability segmentation on the KITTI dataset while attain or slightly surpass the single-task semantic segmentation performance. In a word, the two MTL models degrade the single-task performance (indicated by the negative $\Delta_\texttt{MTL}$ values). Similar observations are also reported by \citet{vandenhende2020multi} on the PASCAL dataset, where the standard MTL baseline underperforms the STL network on tasks such as semantic segmentation, human part segmentation, and saliency detection.

The observed performance degradation can be attributed to the negative transfer issue in which the gradient of the traversability segmentation loss to the shared encoder conflicts with that of the semantic segmentation loss during training. Without any task-specific capabilities, the shared encoder tends to be undertrained for an individual task. This issue is alleviated by adding attention modules to modulate the shared encoder in a task-specific manner: in setting 4, a performance gain is obtained for both semantic segmentation ($+3.43\%$) and traversability segmentation ($+4.11\%$). According to the ablation study, we use our MTL network with task-specific attentions in the following experiments.

\subsection{KITTI Dataset}

KITTI dataset~\cite{geiger2012we} is a real-world benchmark for a variety of computer vision and robotic tasks ranging from visual odometry, object detection and tracking. It is collected by an autonomous driving platform equipped with stereo cameras, a LiDAR scanner, IMU and GPS, etc., in accord with the common setup of a modern robotic system. Recently, KITTI also released its semantic and instance segmentation benchmark. However, none of these benchmarks exactly match this work. Accordingly, we quantitatively evaluate our system using the ground truth data we generated.

\begin{table}[]
    \centering
    \caption{Quantitative results of our system on KITTI dataset using intersection over unions (IoUs) for traversability (trav.) classification.}
    \resizebox{\columnwidth}{!}{
    \begin{tabular}{lccc}
    \toprule
        Method & Untraversable ($\%$) & Traversable ($\%$) & Mean ($\%$) \\
    \midrule
        STL Trav. Segmentation & 92.02 & 75.34 & 83.68 \\
        MTL Trav. Segmentation & 96.42 & 91.42 & 93.92 \\
        Trav. Mapping & 96.81 & 92.34 & 94.58 \\
        Semantic-Trav. Mapping & \bf 97.62 & \bf 94.46 & \bf 96.04 \\
    \bottomrule
    \end{tabular}}
    \label{tab:mlm_kitti}
\end{table}

\begin{figure*}[t]
    \centering
    \hspace{-0.8cm}
    \includegraphics[width=0.6\columnwidth]{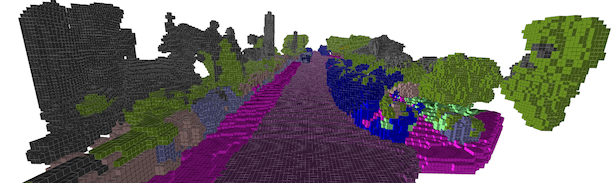}
    \includegraphics[width=0.6\columnwidth]{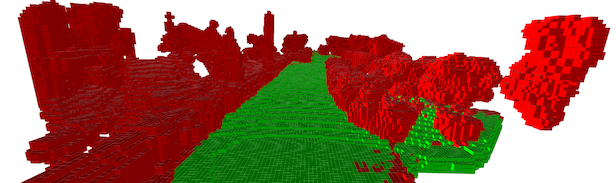}
    \includegraphics[width=0.6\columnwidth]{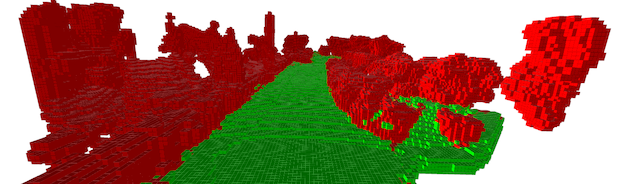} \\
    \vspace{-2.5mm}
    \subfloat[]{\includegraphics[width=0.6\columnwidth]{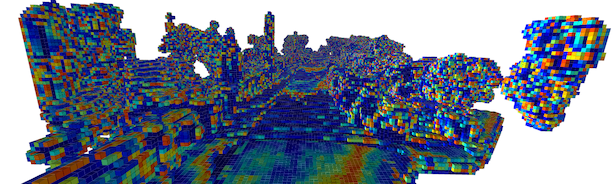}}
    \subfloat[]{\includegraphics[width=0.6\columnwidth]{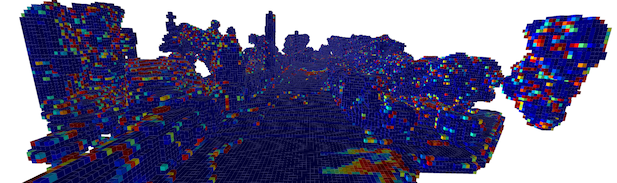}}
    \subfloat[]{\includegraphics[width=0.6\columnwidth]{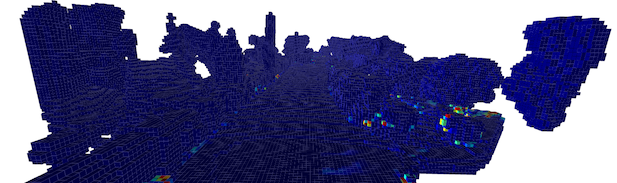}}
    \includegraphics[width=0.07\columnwidth, trim={5cm 2.2cm 4cm 1cm}, clip]{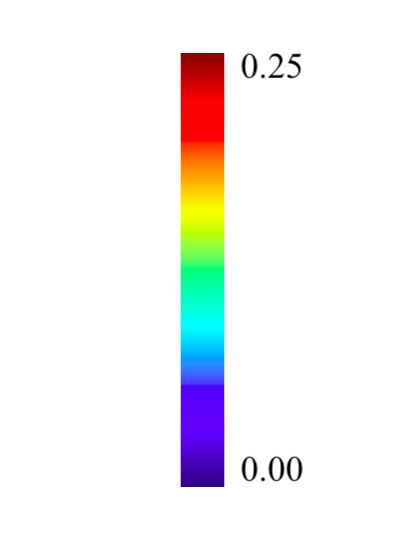}
    \caption{3D qualitative results of our multilayer Bayesian mapping algorithm on KITTI dataset: (a)-(c) are respectively the semantic layer, traversability layer and semantic-traversability layer with its corresponding uncertainty map. The uncertainty map shows the variance of the posterior distribution for each voxel, using Jet colormap for the range of $[0, 0.25]$ (for semantic maps, it shows the variance of the Dirichlet posterior for the predicted class, and for traversability maps, it shows the variance of the beta posterior). We can see the improved estimation results and reduced estimation variances in (c) compared with (b) by leveraging semantic posteriors in traversability inference. Relatively high variances are also observed for the voxels with estimation error.}
    \label{fig:mlm_kitti}
\end{figure*}

\begin{figure*}[t]
    \centering
    \includegraphics[width=0.48\columnwidth]{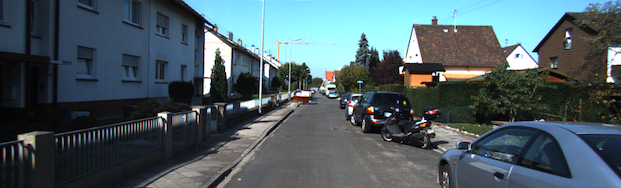}
    \includegraphics[width=0.48\columnwidth]{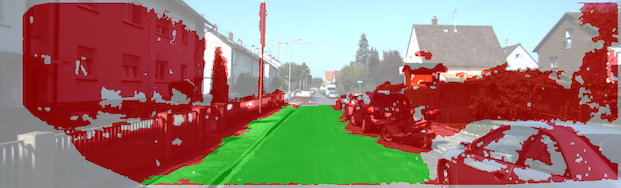}
    \includegraphics[width=0.48\columnwidth]{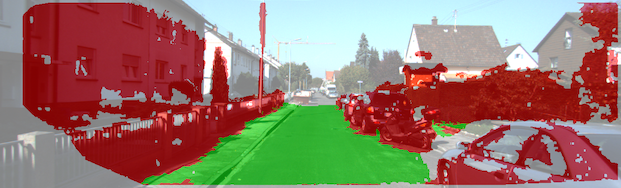}
    \includegraphics[width=0.48\columnwidth]{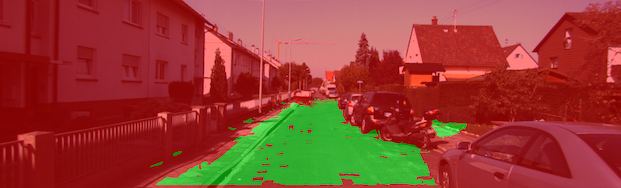} \\
    \vspace{-2.5mm}
    \subfloat[]{\includegraphics[width=0.48\columnwidth]{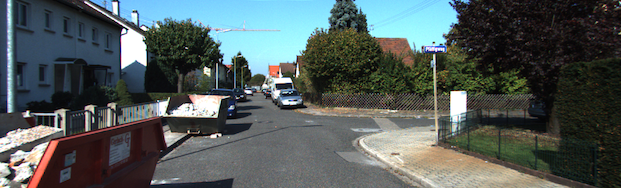}} \hspace{0.1mm}
    \subfloat[]{\includegraphics[width=0.48\columnwidth]{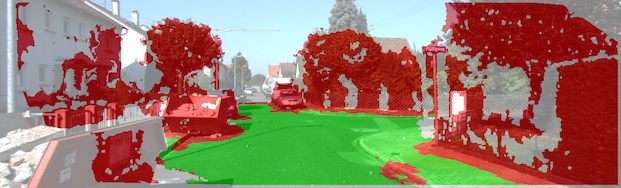}} \hspace{0.1mm}
    \subfloat[]{\includegraphics[width=0.48\columnwidth]{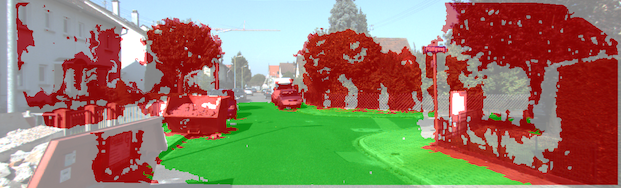}} \hspace{0.1mm}
    \subfloat[]{\includegraphics[width=0.48\columnwidth]{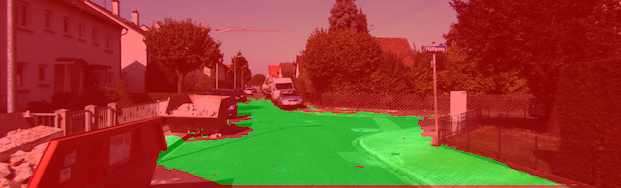}}
    \caption{2D quantitative results of our multilayer Bayesian mapping algorithm on KITTI dataset. (a) Input color images. (b) Projected images from the traversability maps (the pixels with invalid depth are excluded). (c) Projected images from the semantic-traversability maps. (d) Traversability ground truth labels generated automatically using the method in Section~\ref{sec:trav_labeling}. We can see the improved estimation in (c) compared with (b) based on the ground truth.}
    \label{fig:mlm_kitti_reproj}
\end{figure*}

For training our MTL network, we use the KITTI pixel-level semantic segmentation dataset which consists of 200 training images. It shares the same 19 semantic class definition with Cityscapes~\cite{cordts2016cityscapes}, including \emph{road}, \emph{sidewalk}, \emph{building}, \emph{car}, etc. For traversability segmentation, we generate ground truth labels for the KITTI odometry dataset sequence 06 and 07, and randomly select 200 training images for task balancing. We first use ORB-SLAM2~\cite{murORB2} to estimate the 6-DoF camera poses from stereo measurements, and transform them to get the LiDAR poses using camera-LiDAR extrinsics. We then use the LiDAR point clouds and the corresponding LiDAR poses to generate traversability scores, as described in Section~\ref{sec:trav_labeling}. To project the scores onto images, we use the depth values estimated by stereo matching~\cite{Geiger2010ACCV}.

We test our framework on KITTI odometry sequence 15, and provide the quantitative results in Table~\ref{tab:mlm_kitti}. As the ground truth semantic labels for the sequence are unavailable (KITTI does not provide semantic labels for its odometry data; however, we need the sequential information for mapping), we only compare the traversability segmentation performance between the STL baseline and our MTL network. As shown in Table~\ref{tab:mlm_kitti}, our MTL network improves about $10\%$ mIoU for traversability segmentation, and especially improves $16.08\%$ IoU for the traversable class. Two reasons might justify this: First, 200 training images are insufficient to well train an STL network that is pretrained for another task; Second, traversability is highly correlated with semantics in an on-road environment, and the MTL network is more data efficient for those tasks.

The performance of our multilayer Bayesian mapping is also evaluated in Table~\ref{tab:mlm_kitti}, where traversability mapping refers to traversability layer inference without semantic information, and semantic-traversability mapping uses Algorithm~\ref{al:s_t_inference} that leverages inter-layer correlations. For quantitative comparison, we project the 3D maps onto 2D images and compare them with the ground truth images pixel-wisely. To this end, for each pixel in the test image, we use its depth value to query the inferred traversability probability from the corresponding map cell and convert it to a binary prediction by thresholding. We exclude all pixels with invalid depth from both result and ground truth images before computing the metrics. The qualitative results of 3D multilayer maps and 2D projected images are shown in Fig.~\ref{fig:mlm_kitti} and~\ref{fig:mlm_kitti_reproj}, respectively.

The results show that through sequential Bayesian inference, traversability mapping alone can improve the classification accuracy (line 3 compares to line 2 in Table~\ref{tab:mlm_kitti}), which is beneficial from multi-frame measurements. However, leveraging semantic posteriors in traversability inference achieves the best performance (line 4 in Table~\ref{tab:mlm_kitti}). This is also shown in Fig.~\ref{fig:mlm_kitti}, where the semantic-traversability mapping not only corrects some misclassifications, but also reduces the map uncertainty.

\subsection{TartanAir Dataset}

As KITTI dataset aims for on-road autonomous vehicle applications, the environments are semblable and relatively simple. To further test our framework in other environments, we employ a new challenging dataset, the TartanAir~\cite{wang2020tartanair}. TartanAir is mainly introduced as a benchmark for visual SLAM algorithms, but it is also suitable for other robotic applications such as robotic mapping. It is collected in photo-realistic simulation environments and thus able to provide multi-modal sensor measurements (including stereo and depth images, LiDAR point clouds) and precise ground truth data (such as semantic labels and camera poses). 

Unlike Cityscapes or KITTI collected by ground robots from consistent camera viewpoints (pointing to the front from the same height), TartanAir has various camera viewpoints and diverse sensor motion patterns due to its exemption from platform constraints, posing a challenge for the segmentation network. It is also collected in a variety of simulation environments including indoor, outdoor, underwater and sci-fi scenes. Without loss of generality, we choose two representative outdoor environments that are close to the domain of our application for our experiments, i.e., \emph{abandoned factory} and \emph{neighborhood}. For both environments, we use a long sequence for training and several short sequences for testing. Specifically, sequence P000 of 2176 frames is used as training data for the abandoned factory environment, and sequence P000 of 4204 frames for the neighborhood environment.

\begin{table}[]
    \centering
    \caption{Quantitative evaluation of our self-supervised traversability labeling compared with the simulated precise traversability labels on TartanAir training sets using intersection over unions (IoUs).}
    \resizebox{1\columnwidth}{!}{
    \begin{tabular}{lccc}
    \toprule
    Environment & Untraversable ($\%$) & Traversable ($\%$) & Mean($\%$) \\
    \midrule
    Abandoned Factory & 93.42 & 81.80 & 87.61 \\
    Neighborhood & 91.83 & 85.30 & 88.57 \\
    \bottomrule
    \end{tabular}}
    \label{tab:tartanair_traversability_labeling}
\end{table}

\begin{figure*}[t]
    \centering
    \includegraphics[width=0.24\columnwidth]{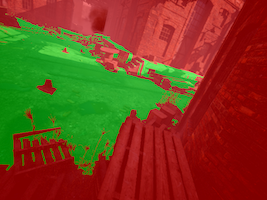}
    \includegraphics[width=0.24\columnwidth]{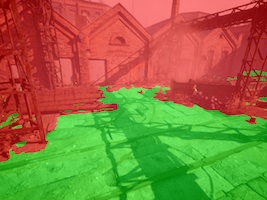}
    \includegraphics[width=0.24\columnwidth]{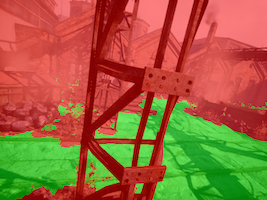}
    \includegraphics[width=0.24\columnwidth]{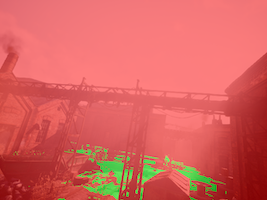}
    \includegraphics[width=0.24\columnwidth]{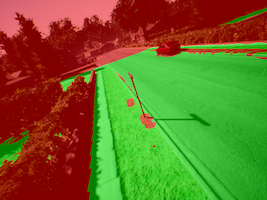} 
    \includegraphics[width=0.24\columnwidth]{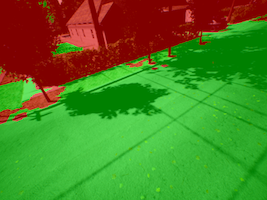}
    \includegraphics[width=0.24\columnwidth]{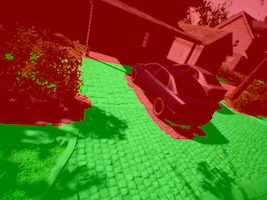}
    \includegraphics[width=0.24\columnwidth]{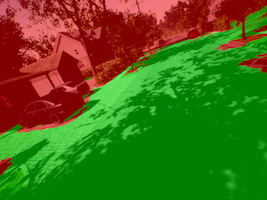}
    \\
    \includegraphics[width=0.24\columnwidth]{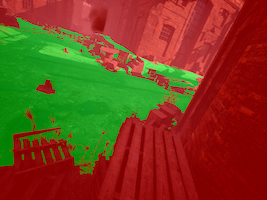}
    \includegraphics[width=0.24\columnwidth]{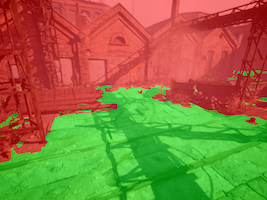}
    \includegraphics[width=0.24\columnwidth]{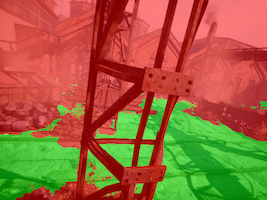}
    \includegraphics[width=0.24\columnwidth]{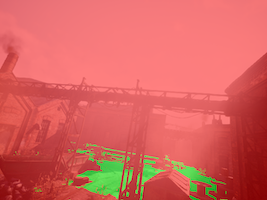}
    \includegraphics[width=0.24\columnwidth]{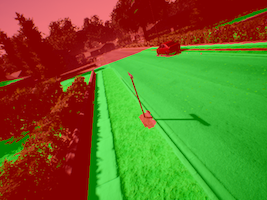}
    \includegraphics[width=0.24\columnwidth]{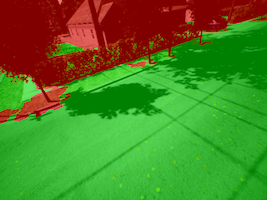}
    \includegraphics[width=0.24\columnwidth]{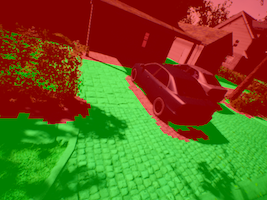}
    \includegraphics[width=0.24\columnwidth]{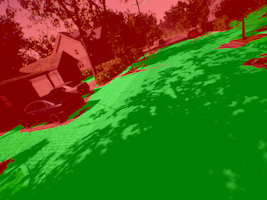}
    \\
    \includegraphics[width=0.24\columnwidth]{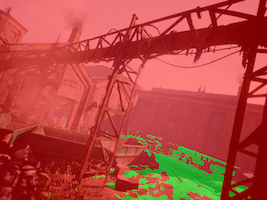}
    \includegraphics[width=0.24\columnwidth]{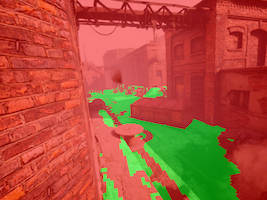}
    \includegraphics[width=0.24\columnwidth]{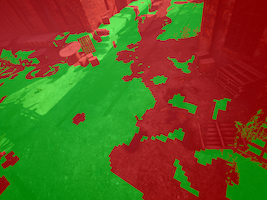}
    \includegraphics[width=0.24\columnwidth]{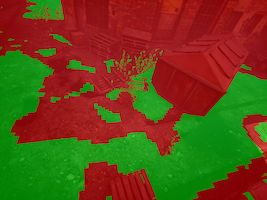}
    \includegraphics[width=0.24\columnwidth]{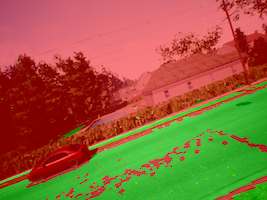} 
    \includegraphics[width=0.24\columnwidth]{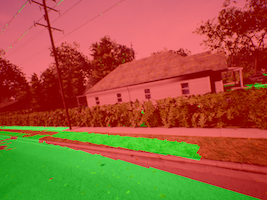}
    \includegraphics[width=0.24\columnwidth]{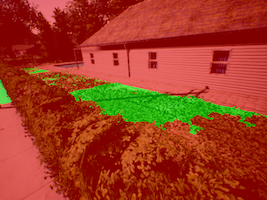}
    \includegraphics[width=0.24\columnwidth]{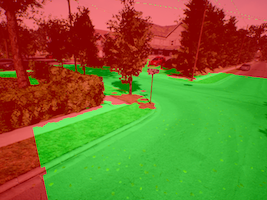}
    \\
    \includegraphics[width=0.24\columnwidth]{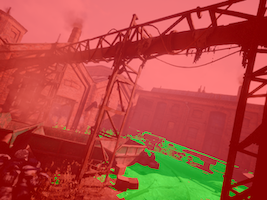}
    \includegraphics[width=0.24\columnwidth]{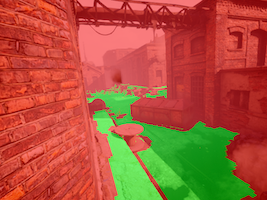}
    \includegraphics[width=0.24\columnwidth]{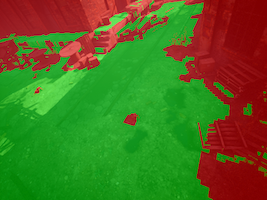}
    \includegraphics[width=0.24\columnwidth]{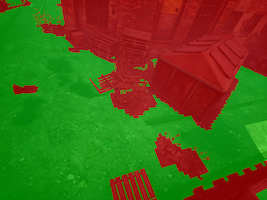}
    \includegraphics[width=0.24\columnwidth]{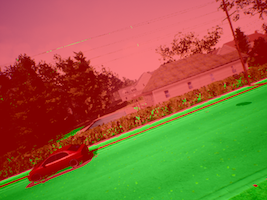} 
    \includegraphics[width=0.24\columnwidth]{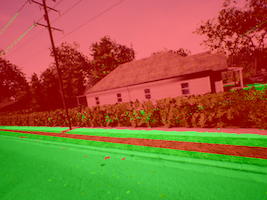}
    \includegraphics[width=0.24\columnwidth]{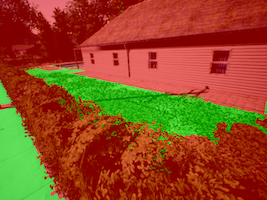}
    \includegraphics[width=0.24\columnwidth]{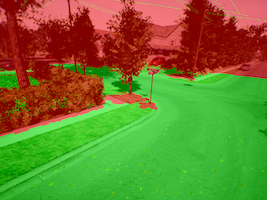}
    \caption{Qualitative evaluation of our self-supervised traversability labeling proposed in Section~\ref{sec:trav_labeling} compared with the simulated precise traversability labels on TartanAir training sets. The left side is generated from sequence P000 of the abandoned factory environment, and the right side is from sequence P000 in the neighborhood environment. The top two rows are the selected best examples of the generated (first row) and simulated precise (second row) traversability labels, and the bottom two rows show the worst cases in the same order. The color images are overlaid with the labels for visualization.}.
    \label{fig:traversability_labeling}
\end{figure*}

\begin{table*}[t]
    \centering
    \caption{Quantitative results of our MTL network on TartanAir test sets. Mean intersection over unions (mIoUs) of 30 semantic classes for the abandoned factory environment, 17 semantic classes for the neighborhood environment, and 2 traversability classes are reported with the multitask performance (${\Delta}_\texttt{MTL}$). The STL and MTL baseline models are defined in setting 1 and 3 in Section~\ref{sec:ablation}.}
    \resizebox{0.85\textwidth}{!}{
    \begin{tabular}{lllccc}
    \toprule
    \bl{Environment} & \bl{Sequence (No. of images)} & Method & Semantic Seg. (mIoU) & Traversability Seg. (mIoU) & ${\Delta}_\texttt{MTL}$ (\%)\\
    \midrule
    \bl{\multirow{12}{*}{Abandoned Factory}} & \bl{\multirow{3}{*}{P001 (434)}} & STL & 47.77 & 85.78 & + 0.00 \\
     &  & \bl{MTL Baseline} & \bl{50.05} & \bl{86.90} & \bl{+ 3.04} \\
     &  & \bl{MTL} & \bl{\bf 53.16} & \bl{\bf 87.01} & \bl{\bf + 6.36} \\
     \cmidrule{2-6}
     & \bl{\multirow{3}{*}{P002 (927)}} & \bl{STL} & \bl{42.61} & \bl{\bf 81.56} & \bl{+ 0.00} \\
     &  & \bl{MTL Baseline} & \bl{42.54} & \bl{79.25}& \bl{- 1.50} \\
     &  & \bl{MTL} & \bl{\bf 46.50} & \bl{80.37} & \bl{\bf + 3.84} \\
     \cmidrule{2-6}
    & \bl{\multirow{3}{*}{P009 (339)}} & \bl{STL} & \bl{44.85} & \bl{82.41} & \bl{+ 0.00} \\
     &  & \bl{MTL Baseline} & \bl{45.33} & \bl{81.78} & \bl{+ 1.53}\\
     &  & \bl{MTL} & \bl{\bf 47.40} & \bl{\bf 84.82} & \bl{\bf + 4.31} \\
    \cmidrule{2-6}
    & \bl{\multirow{3}{*}{Average}} & \bl{STL} & \bl{45.08} & \bl{83.25} & \bl{+ 0.00} \\
     &  & \bl{MTL Baseline} & \bl{45.97} & \bl{82.64} &  \bl{+ 0.62} \\
     &  & \bl{MTL} & \bl{\bf 49.02} & \bl{\bf 84.06} &  \bl{\bf + 4.86} \\
    \midrule
    \bl{\multirow{9}{*}{Neighborhood}} &
    \bl{\multirow{3}{*}{P002 (523)}} & \bl{STL} & \bl{49.74} & \bl{75.36} & \bl{+ 0.00} \\
    & & \bl{MTL Baseline} & \bl{48.92} & \bl{\bf 76.82} & \bl{+ 0.14} \\
    & & \bl{MTL} & \bl{\bf 52.00} & \bl{75.60} & \bl{\bf + 2.43} \\
    \cmidrule{2-6}
    & \bl{\multirow{3}{*}{P005 (724)}} & \bl{STL} & \bl{47.22} & \bl{77.63} & \bl{+ 0.00} \\
    & & \bl{MTL Baseline} & \bl{\bf 51.12} & \bl{77.15} & \bl{\bf + 3.82} \\
    & & \bl{MTL} & \bl{50.37} & \bl{\bf 77.83} & \bl{+ 3.46} \\
    \cmidrule{2-6}
    & \bl{\multirow{3}{*}{Average}} & \bl{STL} & \bl{48.48} & \bl{76.50} & \bl{+ 0.00} \\
    & & \bl{MTL Baseline} & \bl{50.02} & \bl{\bf 76.99} & \bl{+ 1.91} \\
    & & \bl{MTL} & \bl{\bf 51.19} & \bl{76.72} & \bl{\bf + 2.94} \\
    \bottomrule
    \end{tabular}}
    \label{tab:mtl_tartanair}
\end{table*}

\begin{figure*}[t!]
    \centering
    \includegraphics[width=0.28\columnwidth]{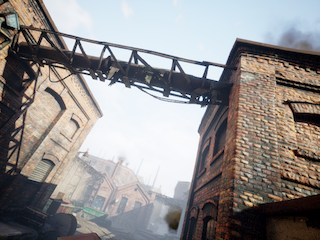}
    \includegraphics[width=0.28\columnwidth]{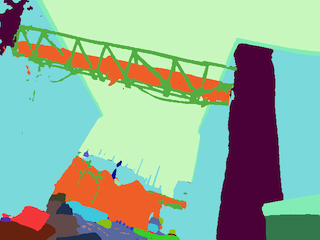}
    \includegraphics[width=0.28\columnwidth]{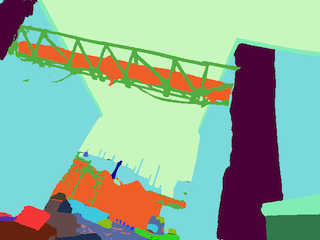} 
    \includegraphics[width=0.28\columnwidth]{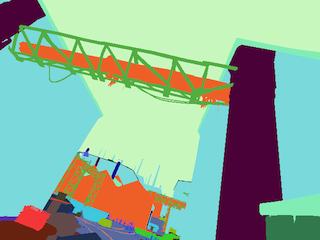}
    \includegraphics[width=0.28\columnwidth]{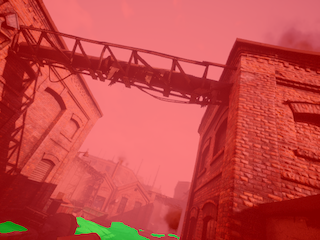}
    \includegraphics[width=0.28\columnwidth]{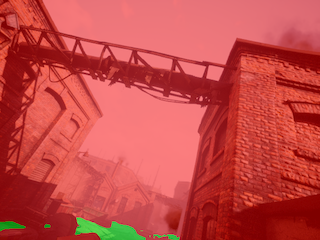}
    \includegraphics[width=0.28\columnwidth]{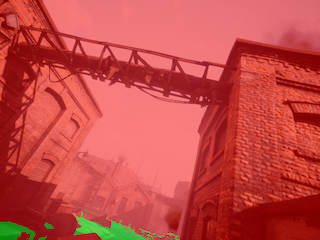} \\
    \includegraphics[width=0.28\columnwidth]{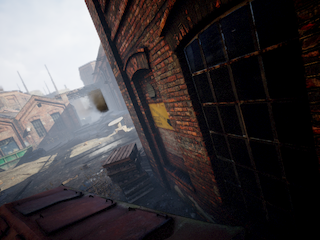}
    \includegraphics[width=0.28\columnwidth]{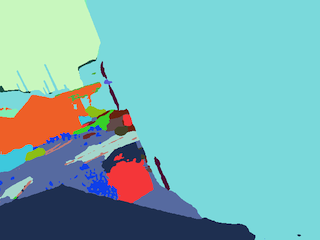}
    \includegraphics[width=0.28\columnwidth]{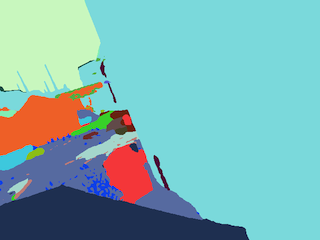} 
    \includegraphics[width=0.28\columnwidth]{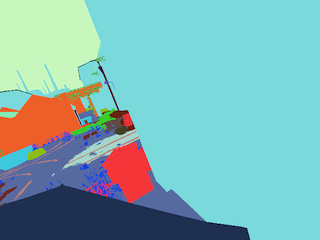}
    \includegraphics[width=0.28\columnwidth]{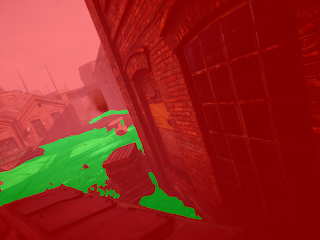}
    \includegraphics[width=0.28\columnwidth]{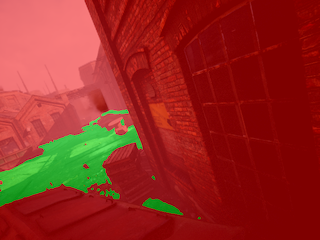}
    \includegraphics[width=0.28\columnwidth]{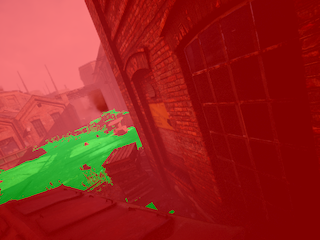} \\
    \includegraphics[width=0.28\columnwidth]{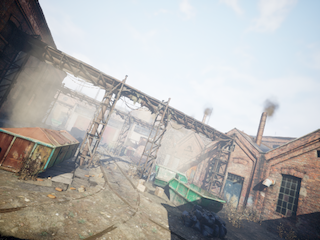}
    \includegraphics[width=0.28\columnwidth]{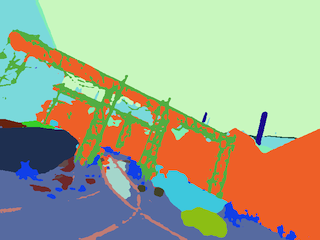}
    \includegraphics[width=0.28\columnwidth]{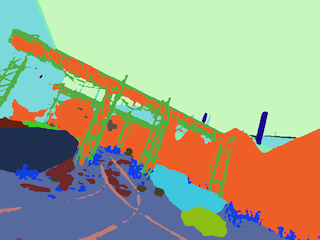} 
    \includegraphics[width=0.28\columnwidth]{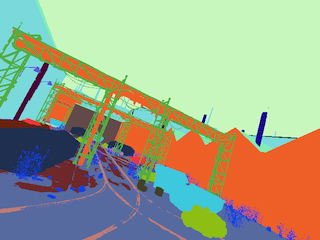}
    \includegraphics[width=0.28\columnwidth]{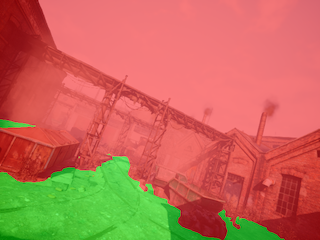}
    \includegraphics[width=0.28\columnwidth]{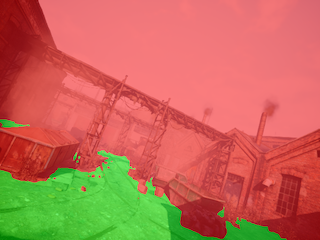}
    \includegraphics[width=0.28\columnwidth]{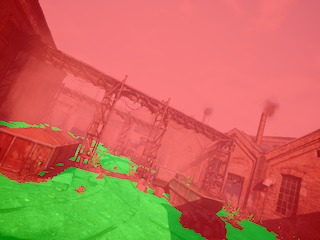} \\
    \includegraphics[width=0.28\columnwidth]{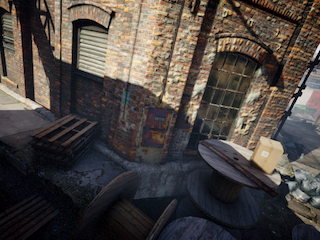}
    \includegraphics[width=0.28\columnwidth]{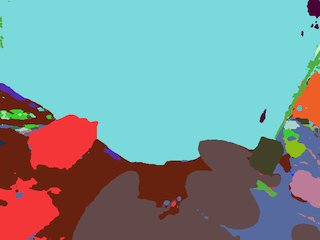}
    \includegraphics[width=0.28\columnwidth]{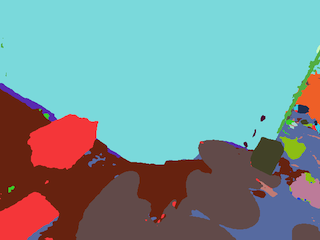} 
    \includegraphics[width=0.28\columnwidth]{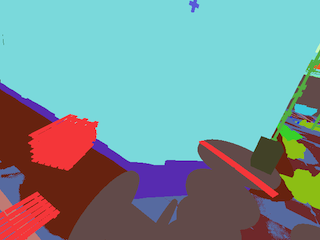}
    \includegraphics[width=0.28\columnwidth]{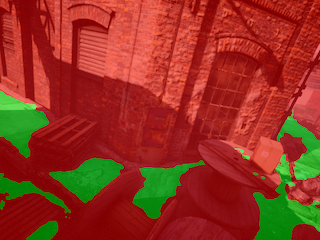}
    \includegraphics[width=0.28\columnwidth]{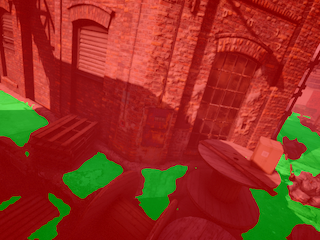}
    \includegraphics[width=0.28\columnwidth]{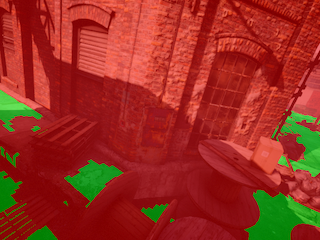} \\
    \includegraphics[width=0.28\columnwidth]{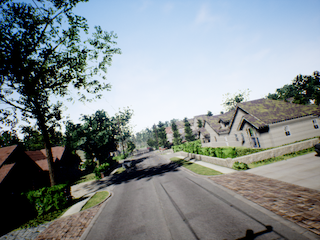}
    \includegraphics[width=0.28\columnwidth]{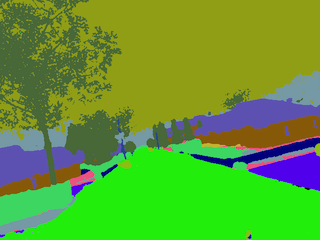}
    \includegraphics[width=0.28\columnwidth]{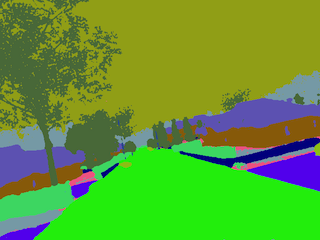}
    \includegraphics[width=0.28\columnwidth]{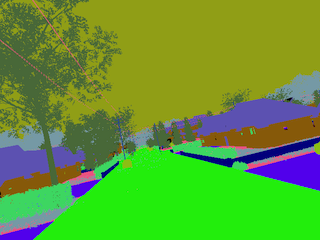}
    \includegraphics[width=0.28\columnwidth]{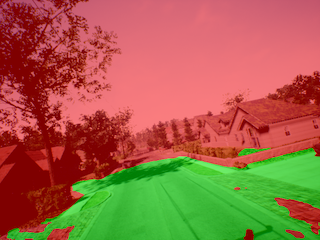}
    \includegraphics[width=0.28\columnwidth]{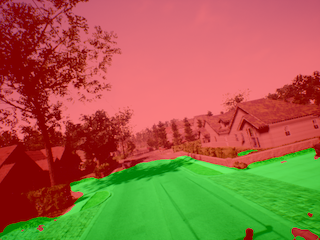}
    \includegraphics[width=0.28\columnwidth]{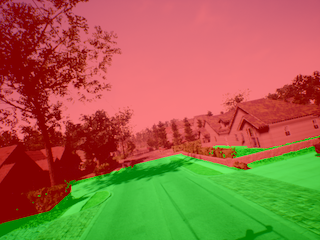} \\
    \includegraphics[width=0.28\columnwidth]{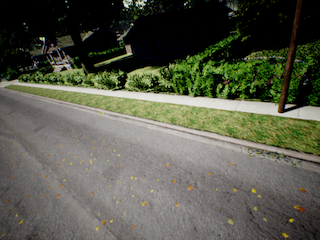}
    \includegraphics[width=0.28\columnwidth]{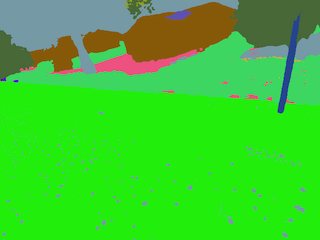}
    \includegraphics[width=0.28\columnwidth]{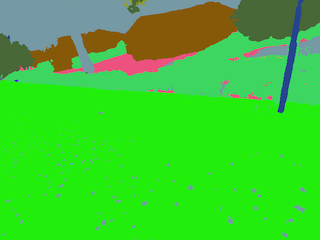} 
    \includegraphics[width=0.28\columnwidth]{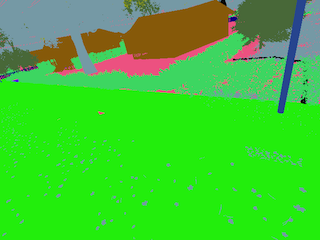}
    \includegraphics[width=0.28\columnwidth]{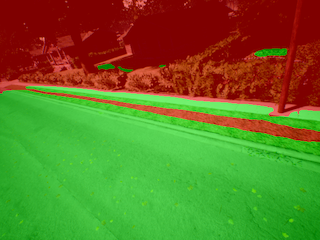}
    \includegraphics[width=0.28\columnwidth]{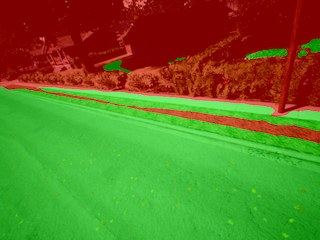}
    \includegraphics[width=0.28\columnwidth]{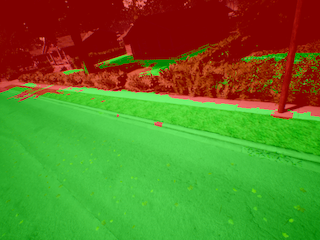}\\
    \includegraphics[width=0.28\columnwidth]{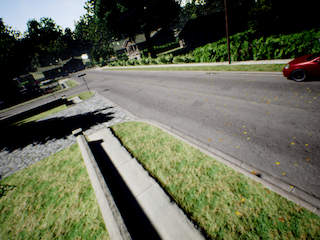}
    \includegraphics[width=0.28\columnwidth]{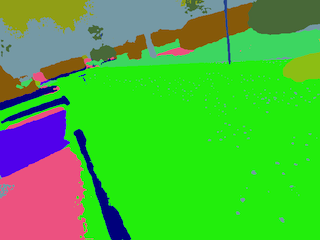}
    \includegraphics[width=0.28\columnwidth]{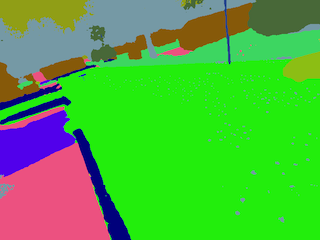} 
    \includegraphics[width=0.28\columnwidth]{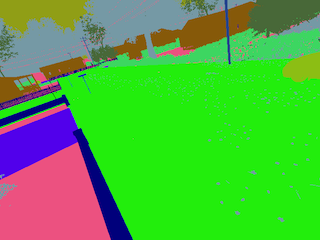}
    \includegraphics[width=0.28\columnwidth]{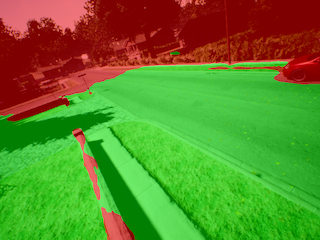}
    \includegraphics[width=0.28\columnwidth]{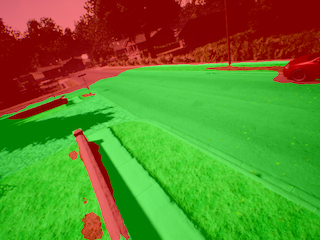}
    \includegraphics[width=0.28\columnwidth]{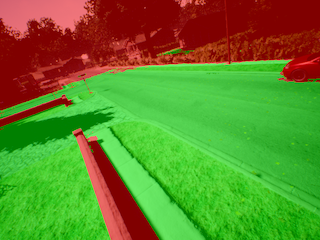} \\
    \vspace{-4mm}
    \subfloat[]{\includegraphics[width=0.28\columnwidth]{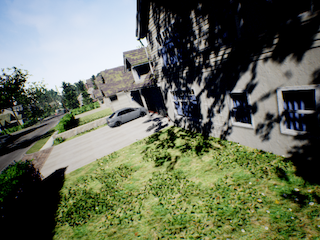}} \hspace{0.2mm}
    \subfloat[]{\includegraphics[width=0.28\columnwidth]{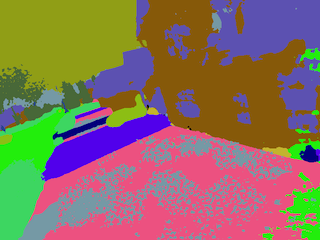}} \hspace{0.2mm}
    \subfloat[]{\includegraphics[width=0.28\columnwidth]{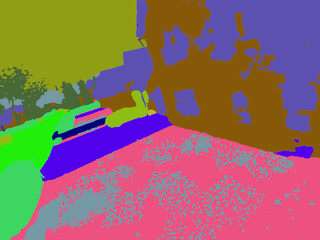}} \hspace{0.2mm}
    \subfloat[]{\includegraphics[width=0.28\columnwidth]{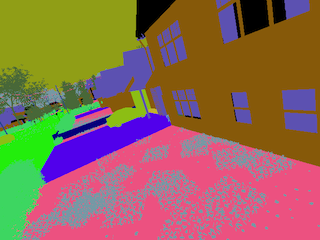}} \hspace{0.2mm}
    \subfloat[]{\includegraphics[width=0.28\columnwidth]{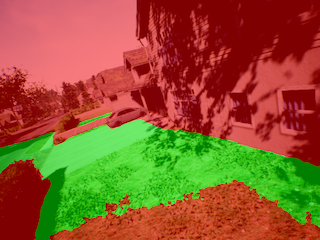}} \hspace{0.2mm}
    \subfloat[]{\includegraphics[width=0.28\columnwidth]{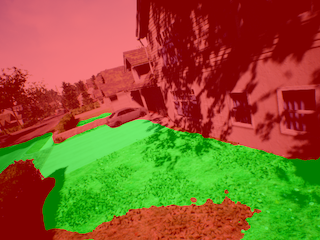}} \hspace{0.2mm}
    \subfloat[]{\includegraphics[width=0.28\columnwidth]{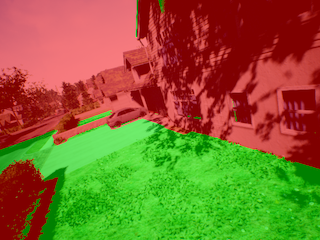}}
    \caption{Qualitative results of our MTL network compared with the STL network on TartanAir dataset. The first four rows are from the abandoned factory environment, and the rest are from the neighborhood environment. For each environment, we show selected examples from the best cases to the worst cases. (a) Input color images. (b)-(d) are STL, MTL, and ground truth semantic segmentation, respectively. (e)-(g) are STL, MTL, and ground truth traversability segmentation, respectively. Note that (g) are automatically generated using the method in~\secref{sec:trav_labeling}. We can see that (c) and (f) have better segmentation results compared with (b) and (e). The performance improvement obtained for the same input image on both tasks also indicates the correlation between the two tasks.}
    \label{fig:mtl_tartanair}
\end{figure*}

\begin{figure*}[t]
    \centering
    \hspace{-0.75cm}
    \includegraphics[width=0.6\columnwidth, trim={38cm 0cm 30cm 0cm}, clip]{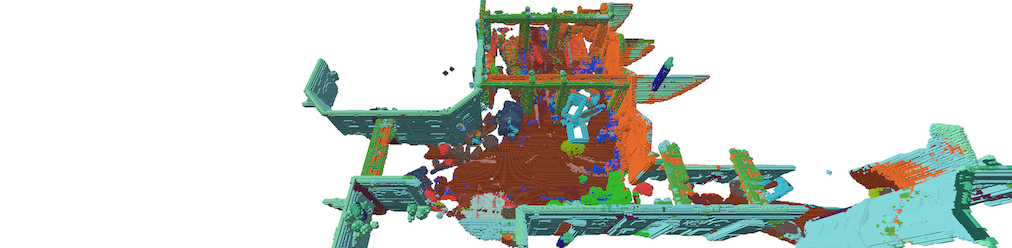}
    \hspace{-0.25cm}
    \includegraphics[width=0.6\columnwidth, trim={38cm 0cm 30cm 0cm}, clip]{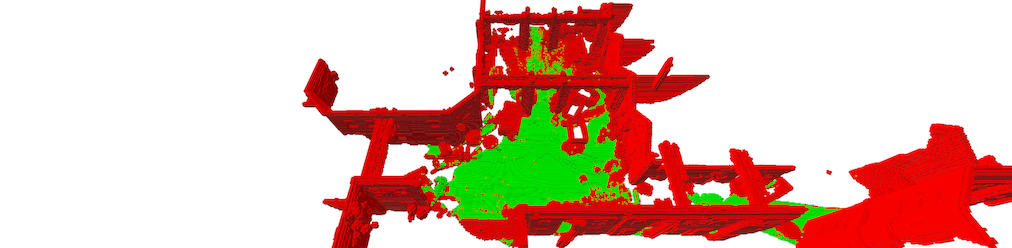}
    \hspace{-0.25cm}
    \includegraphics[width=0.6\columnwidth, trim={38cm 0cm 30cm 0cm}, clip]{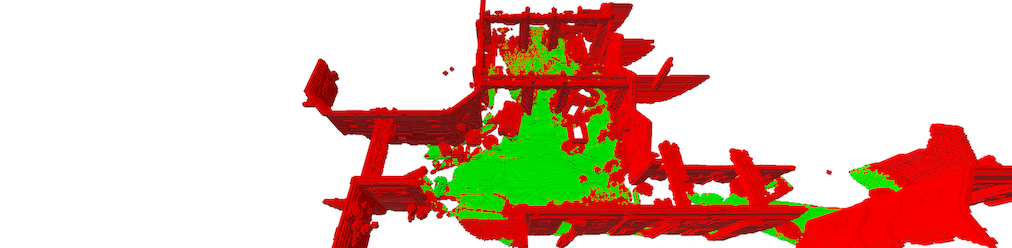} \\
    \vspace{-2.5mm}
    \subfloat[]{\includegraphics[width=0.6\columnwidth, trim={38cm 0cm 30cm 0cm}, clip]{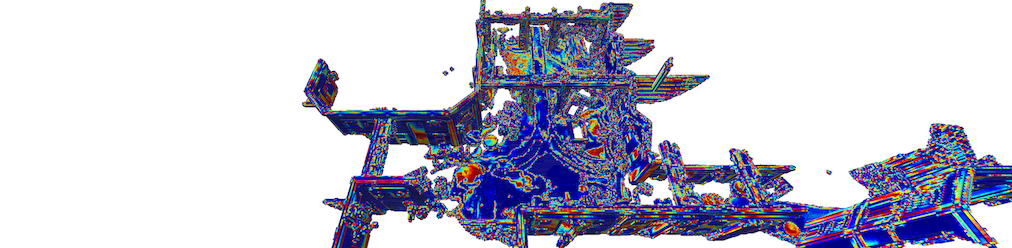}}
    \subfloat[]{\includegraphics[width=0.6\columnwidth, trim={38cm 0cm 30cm 0cm}, clip]{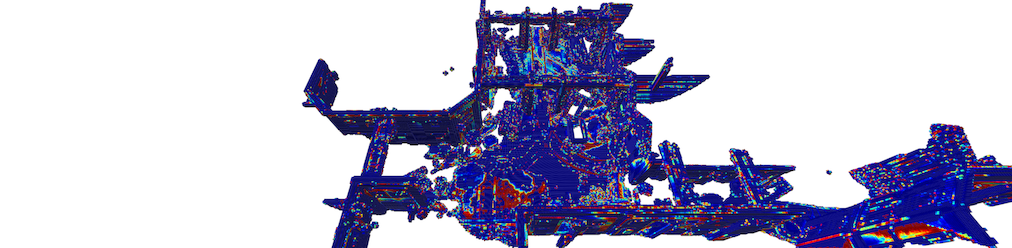}}
    \subfloat[]{\includegraphics[width=0.6\columnwidth, trim={38cm 0cm 30cm 0cm}, clip]{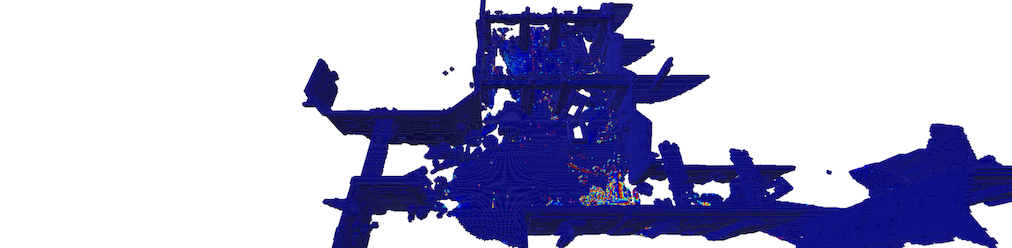}}
    \includegraphics[width=0.08\columnwidth, trim={5cm 2.2cm 4cm 1cm}, clip]{figs/jetcolor_bar_new.png}
    \caption{Qualitative results of our multilayer Bayesian mapping algorithm on TartanAir dataset abandoned factory environment (sequence P001): (a)-(c) are respectively the semantic layer, traversability layer, and semantic-traversability layer with its corresponding uncertainty map. The variance of each voxel is shown using Jet colormap for the range of $[0, 0.25]$.}
    \label{fig:mlm_tartanair}
\end{figure*}

\begin{figure*}[t]
    \centering
    \hspace{-0.8cm}
    \includegraphics[width=0.6\columnwidth, trim={0cm 0cm 0cm 0cm}, clip]{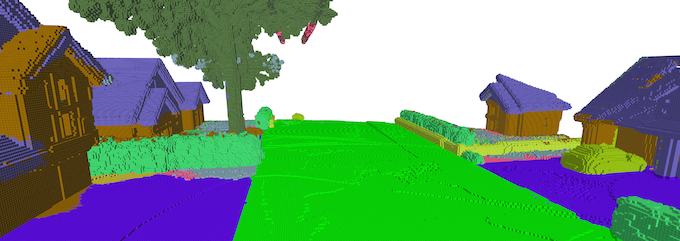}
    \includegraphics[width=0.6\columnwidth, trim={0cm 0cm 0cm 0cm}, clip]{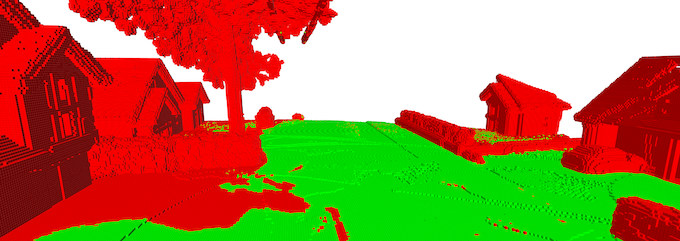}
    \includegraphics[width=0.6\columnwidth, trim={0cm 0cm 0cm 0cm}, clip]{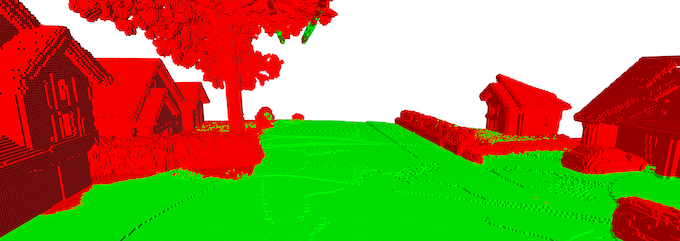}\\
    \vspace{-2.5mm}
    \subfloat[]{\includegraphics[width=0.6\columnwidth, trim={0cm 0cm 0cm 0cm}, clip]{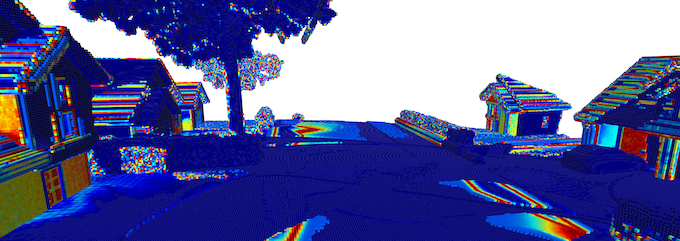}} \hspace{0.2mm}
    \subfloat[]{\includegraphics[width=0.6\columnwidth, trim={0cm 0cm 0cm 0cm}, clip]{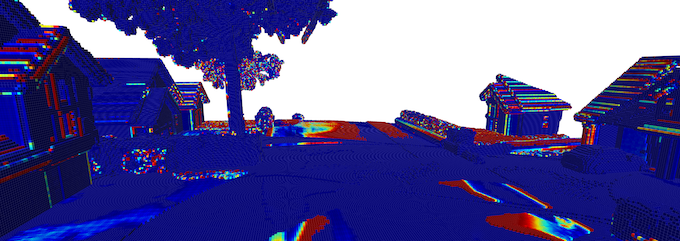}} \hspace{0.2mm}
    \subfloat[]{\includegraphics[width=0.6\columnwidth, trim={0cm 0cm 0cm 0cm}, clip]{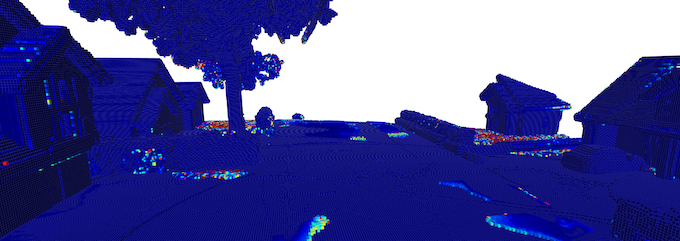}} 
    \includegraphics[width=0.075\columnwidth, trim={5cm 2.2cm 4cm 1cm}, clip]{figs/jetcolor_bar_new.png}
    \caption{Qualitative results of our multilayer Bayesian mapping algorithm on TartanAir dataset neighborhood environment (sequence P005): (a)-(c) are respectively the semantic layer, traversability layer, and semantic-traversability layer with its corresponding uncertainty map. The variance of each voxel is shown using Jet colormap for the range of $[0, 0.25]$.}
    \label{fig:mlm_tartanair_neighborhood}
\end{figure*}

In addition, as the TartanAir dataset provides precise ground truth data for its simulation environments, we can also use these data to evaluate the performance of our self-supervised traversability labeling. To this end, we compare the traversability labels we generated using estimated camera poses and semantic segmentation results with the precise traversability labels generated using the simulated ground truth camera poses (i.e., precise geometry) and semantic labels (i.e., precise semantics) provided by the dataset. The qualitative comparison of both traversability labels is shown in Fig.~\ref{fig:traversability_labeling}, where the upper side shows the best examples and the lower side presents the worst examples. From the worst cases, we can clearly see the errors caused by noisy semantic segmentation (e.g., lower column 1), the drift in pose estimation (e.g., lower column 6) and their combinations. 

The quantitative evaluation is also given in Table~\ref{tab:tartanair_traversability_labeling}. It can be observed that our self-supervised traversability labeling is able to provide reasonably accurate image traversability labels with mean IoUs $87.61\%$ and $88.57\%$ compared with the simulated precise traversability labels. Therefore, we argue that the proposed self-supervised traversability labeling is an effective and inexpensive method to generate image traversability labels for real environments where the geometric and semantic ground truth is usually unavailable. The traversability labeling evaluation is conducted on the training sequences of the TartanAir dataset. In the following experiments, we consistently use the traversability labels generated by our self-supervised method as the traversability ground truth for training and testing.

As TartanAir only provides raw mesh labels randomly assigned by the AirSim simulator, we manually group the mesh labels into semantic labels, resulting in 30 semantic classes for the abandoned factory environment and 17 semantic classes for the neighborhood. Next, we use the semantic labels and our generated traversability labels to train our MTL network. We evaluate the segmentation performance on the test sequences and compare with the corresponding STL network and the standard MTL baseline in~\cite{vandenhende2020multi} (setting 3 in Section~\ref{sec:ablation}). The quantitative results are listed in Table~\ref{tab:mtl_tartanair}.

\begin{table}[]
    \centering
    \caption{Comparison of the average inference time per image per task and the size of model parameters per task among three networks. The experimental setup is given in Section~\ref{sec:implementation}}.
    \resizebox{\columnwidth}{!}{
    \begin{tabular}{lcc}
    \toprule
    Method & Average Inference Time ($\sec$) & Model Parameters (M)\\
    \midrule
    STL & 0.0296 & 137.1 \\
    MTL Baseline & \bf 0.0185 & \bf 85.80 \\
    MTL & 0.0199 & 92.05 \\
    \bottomrule
    \end{tabular}}
    \label{tab:mtl_timing}
\end{table}

We obtain a lower semantic segmentation mIoU on the TartanAir dataset than that on KITTI. This is reasonable as KITTI has a similar environment and the same semantic class definition as Cityscapes on which the network is pretrained. By contrast, TartanAir is more challenging due to the domain gap brought by the differences in environment, class definition and camera viewpoint. In this case, our MTL network still outperforms the STL network in both tasks, and has a better overall performance than the MTL baseline. However, we notice that the mIoU improvement for traversability segmentation is only less than $1\%$. We conjecture that this is because the correlation between the two tasks in TartanAir might not be as high as in KITTI. Some qualitative results of our MTL network are also presented in Fig.~\ref{fig:mtl_tartanair}, where we can see a performance improvement is obtained for both semantic and traversability segmentation compared with single-task learning. The statistics of the inference time and model parameters of three networks are listed in Table~\ref{tab:mtl_timing}. We argue that the main advantage of using an MTL network in our multilayer mapping framework is saving computational resources for robotic applications. Even when the performance improvement of the MTL network is not substantial, STL network does not scale with the number of tasks (map layers).

\begin{table}[]
    \centering
    \caption{Quantitative results of our multilayer Bayesian mapping algorithm on TartanAir dataset using intersection over unions (IoUs) for traversability (trav.) classification.}
    \resizebox{\columnwidth}{!}{
    \begin{tabular}{lccc}
    \toprule
        Method & Untraversable ($\%$) & Traversable ($\%$) & Mean ($\%$) \\
    \midrule
        MTL Trav. Segmentation & 98.63 & 74.71 & 86.67 \\
        Trav. Mapping & 98.81 & 77.43 & 88.12 \\
        Semantic-Trav. Mapping & \bf 99.06 & \bf 80.70 & \bf 89.88 \\
    \bottomrule
    \end{tabular}}
    \label{tab:mlm_tartanair}
\end{table}

The quantitative results of our multilayer Bayesian mapping algorithm are given in Table~\ref{tab:mlm_tartanair}. Following the same trend in KITTI experiments, our traversability mapping improves the classification accuracy of single-frame segmentation by inference on multi-frame measurements (line 2 compares to line 1 in Table~\ref{tab:mlm_tartanair}). Leveraging semantic posterior in traversability inference further improves the mapping performance. It is worth mentioning that our continuous semantic mapping also improves the mIoU of semantic segmentation from $52.30\%$ to $60.30\%$. Although it is not the contribution of this work, our semantic-traversability mapping can benefit from this improvement by using the inferred semantic posterior instead of the noisy semantic segmentation (measurements). Figure~\ref{fig:mlm_tartanair} and~\ref{fig:mlm_tartanair_neighborhood} qualitatively show those map layers with the corresponding uncertainty map for the abandoned factory and neighborhood environment, respectively.

\subsection{Cassie Robot Data}

\begin{figure}[t]
    \centering
    \includegraphics[width=0.98\columnwidth,  trim={0cm 11.5cm 16cm 1.5cm}, clip]{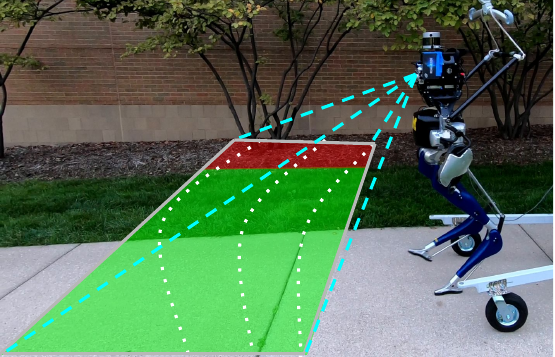}
    \caption{The Cassie Blue robot platform equipped with a depth camera and a LiDAR scanner. We use the LiDAR point clouds and the estimated robot poses to build a 2D local traversability map based on geometric information. We then project the traversability scores onto the color image plane using the depth values from the camera to generate traversability ground truth labels automatically. 
    }
    \label{fig:cassie_robot}
\end{figure}

Lastly, we test our multilayer mapping system on real-world data collected using our Cassie Blue robot platform on the University of Michigan - North Campus. Cassie Blue is a bipedal robot equipped with customized sensor suite including an Intel RealSense depth camera D435 and a Velodyne LiDAR VLP-32C, as shown in Fig.~\ref{fig:cassie_robot}. The extrinsics of camera and LiDAR are calibrated using~\cite{huang2020improvements}. A contact-aided invariant EKF~\cite{hartley2020contact} is running on-board for high-frequent robot pose estimation during data collection. The experiment setup is the same as in~\cite{gan2020bayesian}, except that we build multilayer map offline using recorded data at the recording speed.

For training the MTL network, we extract images and generate traversability labels from a recorded training sequence (ROS bag). We use LiDAR point clouds and the corresponding robot poses from invariant EKF to build a 2D traversability map, and project the traversability scores onto the color image plane using the depth values from the RealSense camera. The self-supervised traversability labeling process is illustrated in Fig.~\ref{fig:cassie_robot}. For semantic labels, we use our manually annotated NCLT dataset collected on campus~\cite{carlevaris2016university}.

Due to the heavy computational requirement (especially GPU memory) of the MTL network, we are currently unable to run the network on the onboard computing device, which is the bottleneck that prevents running our framework fully online. This can be addressed by more efforts on optimizing the network for embedded systems with a performance trade-off, as in our previous work~\cite{gan2020bayesian}. In this experiment, we run the MTL network offline to generate the semantic and traversability segmentation results for the recorded data.

Multilayer mapping algorithm takes the offline semantic and traversability segmentation for map building. It needs to be mentioned that the mapping module is running in an online manner: building the map when data is replayed from a ROS bag. The average runtime of the multilayer mapping is 1.83 $\sec$/scan for both layers.

\begin{figure}
    \centering
    \includegraphics[width=0.48\columnwidth, trim={0cm 2cm 4cm 1.6cm}, clip]{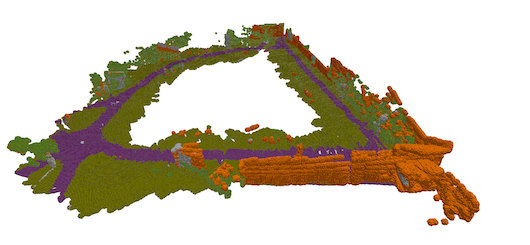}
    \includegraphics[width=0.48\columnwidth, trim={0cm 2cm 4cm 1.6cm}, clip]{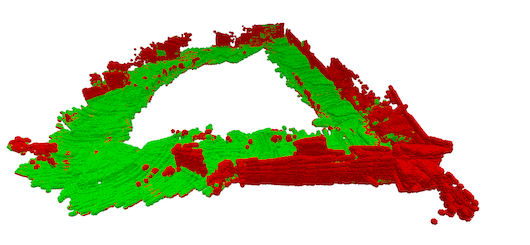} \\
    \includegraphics[width=0.48\columnwidth, trim={0cm 2cm 4cm 1.6cm}, clip]{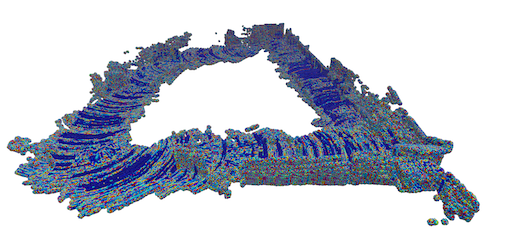}
    \includegraphics[width=0.48\columnwidth, trim={0cm 2cm 4cm 1.6cm}, clip]{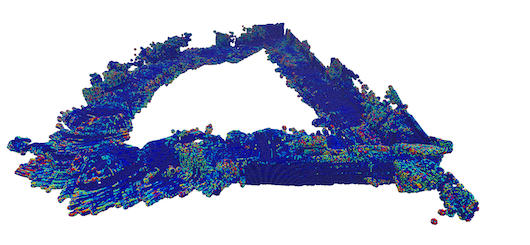} \\
    \vspace{0.15cm}
    \includegraphics[width=0.32\columnwidth, trim={0cm 0cm 0cm 0cm}, clip]{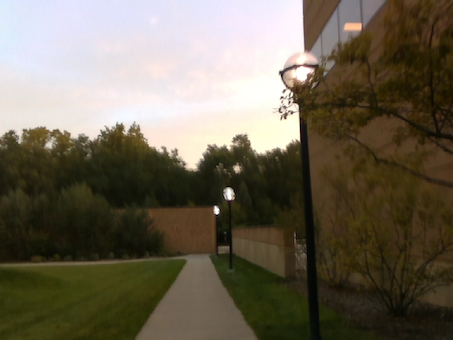}
    \includegraphics[width=0.32\columnwidth, trim={0cm 0cm 0cm 0cm}, clip]{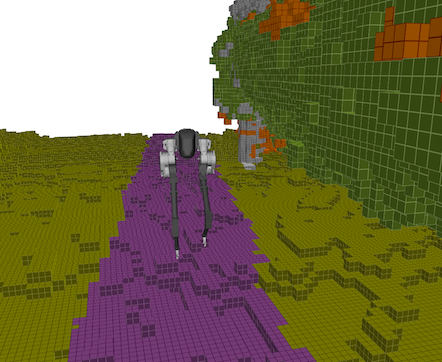}
    \includegraphics[width=0.32\columnwidth, trim={0cm 0cm 0cm 0cm}, clip]{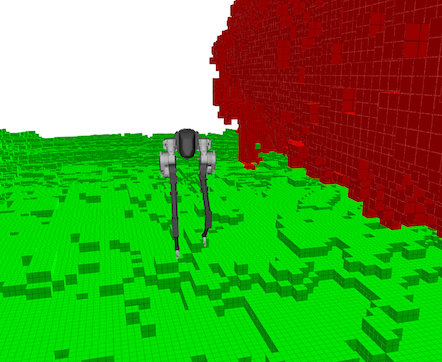}
    \caption{Top block: Qualitative results of our multilayer Bayesian mapping algorithm on data collected with Cassie Blue on the University of Michigan - North Campus. From the left to right column are the semantic layer and semantic-traversability layer with its corresponding uncertainty map. A closed-up view of both map layers are shown in the bottom block with the corresponding input color image captured by the on-board camera shown on the left.}
    \label{fig:mlm_cassie}
\end{figure}

The qualitative results in Fig.~\ref{fig:mlm_cassie} are visually correct and pleasing. Most ground areas are mapped as traversable (green) with some untraversable voxels (red) caused by pedestrians moving in the environment. It is noteworthy that the mapping algorithm processes the recorded data at the original collecting speed, which shows that our multilayer Bayesian mapping is efficient, and ready for online navigation and exploration applications.

\section{Discussion on Multitask Multilayer Mapping Framework}
\label{sec:discussion}
In this section, we discuss the generalizability of the proposed multitask multilayer Bayesian mapping framework and explain why it has the potential of being widely adopted within the robotics community. We first discuss the extension to more map layers via a constructive example. Next, we discuss significant problems that must be addressed for real-time applications of the proposed framework in autonomous navigation. The latter is an indispensable challenge for safe navigation, especially in unknown unstructured environments, and deserves more in-depth studies as future work.

\subsection{Extensions to More Layers}
\label{sec:discussion_extensions}

The current framework can be easily extended to include more map layers. As a constructive example, this subsection discusses a friction coefficient layer and the way to use it in traversability layer inference.

We assume the friction coefficient observation of the $j$-th map cell follows a univariate Gaussian distribution with known variance $\sigma^2$; \mbox{$f_i \sim \mathcal{N} (\mu_j, \sigma^2)$}, where $f_i$ is the $i$-th measurement acquired at 3D position $x_i$ in map cell $j$~\footnote{Of course, obtaining accurate friction coefficient measurements is a highly challenging problem and an interesting future research direction in robotic mapping. Here, we only discuss the mathematical derivation and generalizability of the developed mapping framework.}. Given observations \mbox{$\mathcal{D}_f := \{(x_i, f_i)\}_{i=1}^N$}, we seek the posterior $p(\mu_j|\mathcal{D}_f)$ in friction mapping. A conjugate prior for this Gaussian likelihood function is also a Gaussian distribution \mbox{$\mu_j \sim \mathcal{N}(\mu_0, \sigma_0^2)$}~\cite{bishop2006pattern}. The mean and variance of the posterior distribution are given by:
\begin{align*}
    \nonumber \mathbb{E}[\mu_j | \mathcal{D}_f] &= \frac{\sigma^2}{M\sigma_0^2+\sigma^2}\mu_0 + \frac{\sigma_0^2}{M\sigma_0^2 + \sigma^2} \sum_{i=1}^M f_i,\\
    \mathbb{V}[\mu_j | \mathcal{D}_f] &= \frac{\sigma^2\sigma_0^2}{\sigma^2 + M\sigma_0^2},
\end{align*}
where $M$ is the number of training points in the $j$-th map cell.

To apply Bayesian kernel inference, we can set $\sigma_0^2 = \sigma^2 / \lambda$, where $\lambda$ is a hyperparameter reflecting the confidence in the prior~\cite{shan2018bayesian}. The kernel version of mean and variance are then derived as:
\begin{align*}
    \nonumber \mathbb{E}[\mu_\ast | \mathcal{D}_f] &= \frac{\lambda \mu_0 + \sum_{i = 1}^N k(x_\ast, x_i)f_i}{\lambda + \sum_{i=1}^N k(x_\ast, x_i)},\\
    \mathbb{V}[\mu_\ast | \mathcal{D}_f] &=  \frac{\sigma^2}{\lambda + \sum_{i=1}^N k(x_\ast, x_i)}.
\end{align*}

Similar to the derivation of our semantic traversability Bayesian inference, we can convert the friction coefficient (Gaussian) posterior distribution \mbox{$p(\mu_j | \mathcal{D}_f)$} to a Bernoulli distribution \mbox{$\text{Bernoulli}(p)$} by setting thresholds for the Gaussian cumulative distribution function based on the correlation between friction coefficient and traversability, where \mbox{$p = P(f_{low} \leq \mu_j \leq f_{high})$}. We then sample from this Bernoulli distribution to get friction-traversability measurements \mbox{$\mathcal{F}^\prime := \{f^\prime_1, ..., f^\prime_N | f^\prime_i \in \{0, 1\}\}$} as done in Algorithm~\ref{al:s_t_inference}.

To incorporate fiction coefficient layer inference into the traversability mapping, we assume the measurements $\mathcal{F}^{\prime}$ are independent to $\mathcal{Z}$ and $\mathcal{Y}^\prime$, and also have Bernoulli likelihood:
\begin{equation*}
    p(f^\prime_i | \phi_j ) = \phi_j^{f^{\prime}_i}(1-\phi_j)^{1-f^{\prime}_i},
\end{equation*}
where $\phi_j, \mathcal{Z}$ and $\mathcal{Y}^\prime$ are defined in Section~\ref{sec:s_t_mapping}. Following the rest derivations in Section~\ref{sec:s_t_mapping}, we are able to leverage the correlations to both semantic layer and fiction coefficient layer in traversability mapping. The traversability posterior \mbox{$p(\phi_j|\mathcal{D}_{y^\prime}, \mathcal{D}_{f^\prime}, \mathcal{D}_{z})$} can then be obtained as $\text{Beta}(\alpha_j, \beta_j)$ with:
\begin{align*}
    \alpha_j &:= \alpha_0 + \sum_{i=1}^N k(x_j, x_i) (y^\prime_i + f^\prime_i + z_i), \\
    \beta_j &:= \beta_0 + \sum_{i=1}^N k(x_j, x_i) (3- y^\prime_i - f^\prime_i - z_i).
\end{align*}
The mean and variance of $\phi_j$ remain the same as in \eqref{eq:variance}.

\begin{remark}
 We note that the provided example does not correspond to the best approach for modeling the stated problem. It merely serves as a constructive example of adding more map layers. The general Bayesian inference~\cite{bishop2006pattern}, of course, can produce more accurate results; however, the motivation for the example is to maintain the closed-form Bayesian inference of each map layer.
\end{remark}

\subsection{Limitations}
There are several limitations in the current work. The current Bayesian map inference is based on a static-world assumption, i.e., $p(\mathcal{M}_t) = p(\mathcal{M}_{t-1})$, where $t$ indicates the measurement timestamp. To deal with dynamic objects and environmental change, a prediction step $p(\mathcal{M}_t | \mathcal{M}_{t-1}, \mathcal{D})$ needs to be modeled given measurements from change detection or scene flow estimation. The work of~\cite{unnikrishnan2021dynamic} is an extension of \cite{gan2020bayesian} to dynamic environments using a simple autoregressive transition model for scene propagation in closed form. For a recent literature review on dynamic semantic mapping and scene flow estimation, we refer to \cite{unnikrishnan2021dynamic} and references therein.

In addition, the current traversability notion is robot-agnostic, as we only use exteroceptive sensor data as the supervisory signal. For navigation and exploration purposes, a \emph{dynamic cost} map layer could be learned from the robot proprioceptive or multi-modal sensory information, where the cost reflects dynamic traversability based on the current operating point and robot behavior~\cite{gan2022energy}. The framework developed in this work provides the foundation for implementing these ideas in the future.

\subsection{Reproducibility}

The current results and maps are reproducible using the same datasets for two reasons. First, the traversability ground truth labels are generated using the provided data without any manual labeling. Second, the multilayer Bayesian map inference provides exact solutions without approximation. This framework can also be used on customized data collected using other robotic platforms as long as the image and depth information are included.

\section{Conclusion}
\label{sec:conclusion}

This paper developed a multitask multilayer Bayesian mapping framework that uses a deep MTL network as a unified reasoning block. The proposed framework 
\begin{enumerate}
    \item provides multiple high-level measurements simultaneously,
    \item learns map attributes other than semantics in a self-supervised manner, and
    \item infers a multilayer dense map in closed form where inter-layer correlations are leveraged.
\end{enumerate}
As a constructive example and a useful case, we specifically build a robotic map with semantic and traversability layers. Experimental results on publicly available datasets and data collected by our robot platform show the advantage of using an MTL network for multilayer mapping, and the performance improvement of traversability inference when the correlation with semantic layer is incorporated.

The proposed framework is highly extendable to include additional map layers containing greater detail, such as friction coefficient, affordance, dynamic planning cost, etc., to assist more sophisticated robotic behavior planning. The map is also continuous and contains uncertainty information, which is desirable for decision-making problems. We hope that the proposed mapping framework is adopted and extended to fulfill many advanced real-world robotic applications.

\ifCLASSOPTIONcaptionsoff
  \newpage
\fi

{\small
\bibliographystyle{template/IEEEtranN}
\bibliography{bib/strings-abrv, bib/IEEEabrv, bib/refs}
}

\end{document}